\documentclass[10pt,twocolumn,letterpaper]{article}

\usepackage{cvpr}              %
\usepackage[accsupp]{axessibility} %

\usepackage[normalem]{ulem}
\usepackage[dvipsnames]{xcolor}

\definecolor{cvprblue}{rgb}{0.21,0.49,0.74}
\usepackage[pagebackref,breaklinks,colorlinks,citecolor=cvprblue]{hyperref}

\def\paperID{15300} %
\def\confName{CVPR}
\def\confYear{2024}

\usepackage{makecell}
\usepackage{tabulary}
\usepackage{multirow}
\usepackage{colortbl}
\usepackage{fontawesome5}

\definecolor{demphcolor}{RGB}{90,90,90}
\newcommand{\demph}[1]{\textcolor{demphcolor}{#1}}
\usepackage{pifont}%
\newcommand{\cmark}{\color{mygray}\ding{51}}%
\newlength\savewidth\newcommand\shline{\noalign{\global\savewidth\arrayrulewidth
  \global\arrayrulewidth 1pt}\hline\noalign{\global\arrayrulewidth\savewidth}}

\newcommand{\tablestyle}[2]{\setlength{\tabcolsep}{#1}\renewcommand{\arraystretch}{#2}\centering\footnotesize}
\usepackage{xspace}

\newcommand{\MP}{\textit{Multi-Path}\xspace}
\newcommand{\method}{\textit{Troika}\xspace}
\newcommand{\CMT}{\textit{Cross-Modal Traction}\xspace}
\definecolor{mygray}{gray}{0.4}

\definecolor{rightcolor}{RGB}{0,144,81}
\definecolor{wrongcolor}{RGB}{255,38,0}

\title{Troika: Multi-Path Cross-Modal Traction for Compositional Zero-Shot Learning}

\author{Siteng Huang$^{1, 3}$\thanks{Work done during internship at Alibaba Group.}, Biao Gong$^{2}$, Yutong Feng$^{2}$, Min Zhang$^{3}$, Yiliang Lv$^{2}$, Donglin Wang$^{3}$\thanks{Corresponding author.}\\
{$^1$Zhejiang University}\ \ {$^2$Alibaba Group}\\
{$^3$Machine Intelligence Lab (MiLAB), AI Division, School of Engineering, Westlake University}\\
{\tt\small \{siteng.huang, a.biao.gong\}@gmail.com}\\[-3pt]
{\tt\small \{zhangmin, wangdonglin\}@westlake.edu.cn,}, {\tt\small \{fengyutong.fyt, yiliang.lyl\}@alibaba-inc.com}
}

\begin{document}
\maketitle

\begin{abstract}
   Recent compositional zero-shot learning (CZSL) methods adapt pre-trained vision-language models (VLMs) by constructing trainable prompts only for composed state-object pairs.
   Relying on learning the joint representation of seen compositions, these methods ignore the explicit modeling of the state and object, thus limiting the exploitation of pre-trained knowledge and generalization to unseen compositions.
   With a particular focus on the universality of the solution,
   in this work, we propose a novel paradigm for CZSL models that establishes three identification branches (\textit{i.e.}, \MP) to jointly model the state, object, and composition.
   The presented \textbf{\method} is an outstanding implementation that aligns the branch-specific prompt representations with decomposed visual features.
   To calibrate the bias between semantically similar multi-modal representations, we further devise a \CMT module into \textbf{\method} that shifts the prompt representation towards the current visual content.
   We conduct extensive experiments on three popular benchmarks, where our method significantly outperforms existing methods in both closed-world and open-world settings.
   The code will be available at \url{https://github.com/bighuang624/Troika}.
\end{abstract}

\section{Introduction} \label{sec:introduction}

As for the study of human-like compositional generalization ability, compositional zero-shot learning (CZSL)~\cite{Misra:red-wine-to-red-tomato,Li:symmetry-and-group,Purushwalkam:task-driven-modular-networks} studies to recognize unseen \textit{compositions} at test time, while states and objects (\textit{i.e.}, \textit{primitives}) are presented in seen \textit{compositions} during training.
Rather than learning to associate images with such state-object compositions from scratch~\cite{Naeem:CGE,Mancini:Co-CGE}, recent efforts~\cite{Nayak:CSP,Xu:PromptCompVL,Lu:DFSP} focus on adapting pre-trained vision-language models (VLMs), \textit{e.g.}, CLIP~\cite{Radford:CLIP}. 
Since CZSL datasets provide only compositional labels (\textit{e.g.}, ``\textit{red}''+``\textit{wine}'') instead of complete sentences (\textit{e.g.}, ``\textit{A glass of wine placed on the table}'') used in pre-training, to bridge the gap without fine-tuning the entire model, prompts~\cite{Lester:prompt-tuning,Liu:prompt-tuning} are constructed by appending a simple prefix like ``\textit{a photo of}'' before the composed state-object labels.
Prior methods~\cite{Nayak:CSP,Xu:PromptCompVL,Lu:DFSP} have replaced the fixed prompt tokens with learnable tokens that are directly optimized during fine-tuning.
By designing prompt tuning solutions for compositions, existing methods can efficiently contrive a cross-modal alignment between images and compositional labels, thereby unlocking the potential compositional generalization capability of VLMs.

\begin{figure}[!t]
  \centering
   \includegraphics[width=\linewidth]{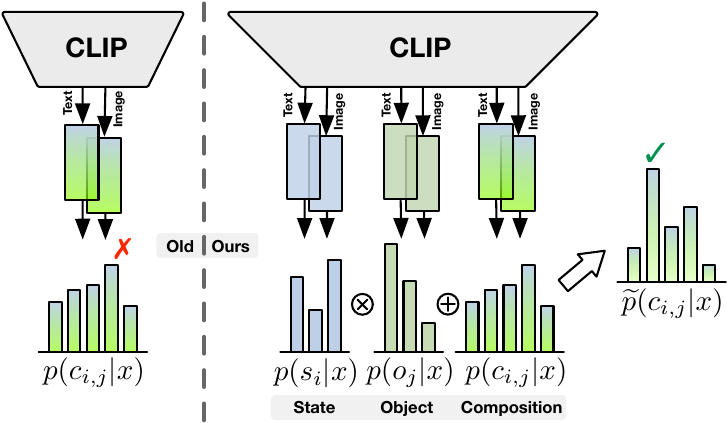}
   \vspace{-7mm}
   \caption{
   \textbf{Graphical comparison of the existing paradigm and the proposed \MP paradigm.}
   }
   \label{fig:multi-path}
   \vspace{-5mm}
\end{figure}

However, due to the lack of independent and explicit \textit{primitive modeling}, these methods have suffered from two challenges:
\textbf{(1)} the full leveraging of pre-trained knowledge fails, since a large amount of cross-modal information is not tied to the compositions, but related to the single primitive.
\textbf{(2)} the difficulty of generalizing to unseen compositions is increased, since the model easily over-rely on a limited number of seen compositions.
To overcome the issues with a particular focus on the universality of the solution, 
we propose a novel \textbf{\MP} paradigm for CZSL with VLMs. 
As shown in \cref{fig:multi-path}, our paradigm emphasizes the joint modeling of the state, object, and composition without redundant assumptions about the specific implementations.
Different from previous methods that depend only on the estimated composition probability, 
\MP paradigm integrates the predictions of all semantic components for the final decision.
Following the accessible paradigm, we present an outstanding implementation \textbf{\method}\footnote{``Troika'' is a traditional harness driving combination, using three horses abreast, usually pulling a sleigh.}.
On the language side, 
\method constructs branch-specific prompts that inject learnable priors for describing the context of specific target classes.
And on the vision side, 
while introducing parameter-efficient adaptation,
\method decomposes primitive visual features for individual recognition.

\begin{figure}[!t]
  \centering
   \includegraphics[width=\linewidth]{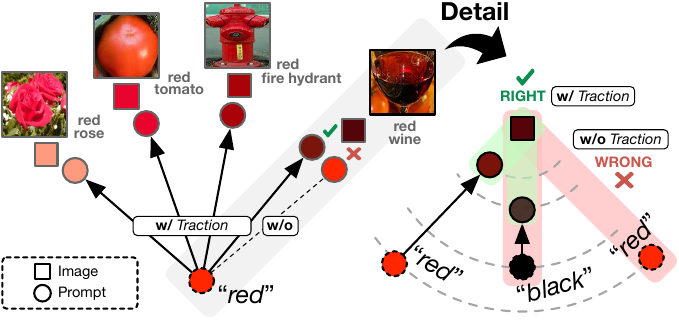}
   \vspace{-7mm}
   \caption{
   \textbf{An example of the \CMT module.}
   The commonly learned prompt of ``\textit{red}'' may be further away (compared to ``\textit{black}'') from individual images with the same concept, and
   the module reduces such mismatches by adaptively pulling the prompt representation towards the current visual content.
   }
   \label{fig:CMT}
   \vspace{-5mm}
\end{figure}

Moreover, for calibrating the bias between semantically similar multi-modal representations, we further devise a \textbf{\CMT} module into \method.
The motivation is that compared to diverse visual presentations, learning only a fixed prompt representation is intuitively insufficient to match all corresponding images from different domains (\cref{fig:CMT}).
By selecting and integrating the most semantically relevant visual features, the module pulls the originally static prompt representation towards the visual content.
As shown in \cref{tab:ablation-fusion}, while the basic \method has already achieved state-of-the-art (SOTA), the incorporation of the \CMT module leads to a significant improvement.
Follow-up researches are also free to try other traction ways as the module and \method are decoupled.

Three popular benchmark datasets MIT-States~\cite{Isola:MIT-States}, UT-Zappos~\cite{Yu:UT-Zappos}, and C-GQA~\cite{Naeem:CGE} are used for comparisons.
Experiments show that on the closed-world setting, \method exceeds the current state-of-the-art methods by up to \textbf{+7.4\%} HM and \textbf{+5.7\%} AUC.
And on the more challenging open-world setting, \method still surpasses the best CLIP-based method by up to \textbf{+3.8\%} HM and \textbf{+2.7\%} AUC.
We also conduct abundant ablations to verify the effectiveness of all component elements of \method.
In summary, the main contributions of our work are four-fold:
\begin{itemize}[labelsep=0.4em, leftmargin=1em,itemindent=0em]
	\item We propose a novel \MP paradigm for CZSL with VLMs, which explicitly constructs vision-language alignments for the state, object, and composition.
    The paradigm is flexible enough to derive new approaches.
	\item Based on the paradigm, we implement a model named \method that effectively aligns the branch-specific prompt representations and decomposed visual features.
    \item We further design a \CMT module for \method that adaptively adjusts the prompt representation depending on the visual content.
    \item We conduct extensive experiments on three CZSL benchmark datasets to show that \method achieves the SOTA performance on both closed-world and open-world settings.
\end{itemize}

\section{Related Work}

\paragraph{Compositional Zero-Shot Learning (CZSL).}
Aiming to recognize 
unseen state-object pairs at test time while each semantic primitive exists in training samples, early CZSL efforts can be broadly divided into two lines.
One line of works~\cite{Misra:red-wine-to-red-tomato,Purushwalkam:task-driven-modular-networks,Naeem:CGE,Mancini:Co-CGE,Anwaar:CVGAE} learns the combined state-object semantic representation for both seen and unseen compositions with a transformation function, \textit{e.g.}, a multi-layer perceptron (MLP)~\cite{Misra:red-wine-to-red-tomato} or a graph convolutional network~\cite{Naeem:CGE}.
Another line of works~\cite{Li:symmetry-and-group,Mancini:CompCos,Karthik:KG-SP,Li:SCEN,Tian:IVR} learns two individual classifiers to identify state and object separately from the image features.
While both lines build the connection between visual features and compositional labels from scratch,
recent works focus on transferring the encyclopedic knowledge from pre-trained VLMs.
CSP~\cite{Nayak:CSP} first 
adapts
the CLIP model~\cite{Radford:CLIP} by replacing the classes in textual prompts with trainable state and object tokens.
PromptCompVL~\cite{Xu:PromptCompVL} creates a fully learnable soft prompt including the prefix, state, and object.
The latest DFSP~\cite{Lu:DFSP} proposes a cross-modal decomposed fusion module to learn more expressive image features.
In this work, by proposing the \MP paradigm for CZSL with VLMs, we highlight the importance of jointly and explicitly modeling the state, object, and composition.

\begin{figure}[!t]
  \centering
   \includegraphics[width=\linewidth]{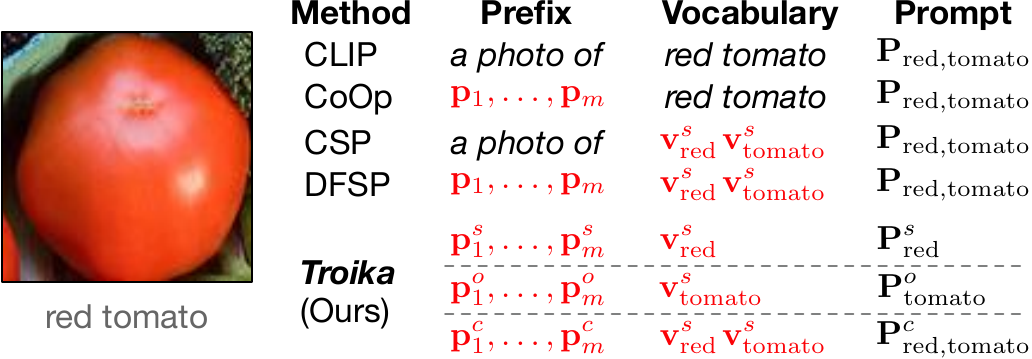}
   \vspace{-7mm}
   \caption{
   \textbf{Graphical comparison of prompts constructed by prior methods and \method.}
   \textcolor{red}{Red} tokens are trainable.
   }
   \label{fig:prompt_comparison}
   \vspace{-5mm}
\end{figure}

\vspace{-5mm}
\paragraph{Parameter-Efficient Transfer Learning (PETL).}
PETL refers to updating only a small number of pre-trained or additional parameters during fine-tuning~\cite{He:unify-PET,Ding:delta-tuning}, 
which reduces the training and storage burdens.
As a popular PETL technique,
\textit{prompt tuning}~\cite{Liu:prompt-tuning,Lester:prompt-tuning,Liu:P-Tuning-deep} optimizes learnable tokens inserted into the input token sequence while freezing the backbone.
CLIP-based CZSL methods~\cite{Nayak:CSP,Xu:PromptCompVL,Lu:DFSP} continue the vein by tuning both the inserted prefix and the primitive vocabulary tokens on downstream semantics.
While following the \MP paradigm to establish three independent branches, 
our \method constructs branch-specific prompts with individual prefixes and a shared primitive vocabulary.
\cref{fig:prompt_comparison} illustrates the differences in prompt design between \method and existing methods, and experimental results in \cref{sec:ablation-study} show that our design benefits from modeling prior knowledge specified with semantic roles.

In this work, we also attempt to introduce PETL techniques to the vision side, implemented as Adapter~\cite{Houlsby:Adapter}.
Since only tuning the text side while ignoring the image encoder is naturally considered insufficient, the studies of adapting both encoders for multi-modal tasks including image recognition~\cite{Zang:UPT,Khattak:MaPLe} and video-text retrieval~\cite{Jiang:Cross-Modal-Adapter,Huang:VoP} have recently emerged.
On these tasks, completely adapting both encoders has been proved to mutually promote the alignment of vision-language modalities.
For the first time, we empirically demonstrate that it is equally valid for CZSL.

\section{Rethinking the Paradigm}

In this section, we first formalize the CZSL task and the visual feature extraction of CLIP~\cite{Radford:CLIP} (\cref{sec:preliminaries}).
Then, we introduce how previous works adapt CLIP by adjusting prompts for image-composition alignments (\cref{sec:framework-revisit}).
Finally, we present a \MP paradigm to guide the construction of VLM-based pipelines (\cref{sec:multi-path-paradigm}).

\subsection{Preliminaries} \label{sec:preliminaries}

\begin{table}[!t]
\tablestyle{5pt}{1.0}
\setlength\tabcolsep{4pt}
\def\w{20pt} 
\scalebox{1}{
    \begin{tabular}{l|cccc|cccc}
          & \multicolumn{4}{c|}{\textbf{CLIP}~\cite{Radford:CLIP}}     & \multicolumn{4}{c}{\textbf{CoOp}~\cite{Zhou:CoOp}} \\
          & S     & U     & HM    & AUC   & S     & U     & HM    & AUC \\
    \shline
    w/o MP & 15.8  & 49.1  & 15.6  & 5.0   & 52.1  & 49.3  & 34.6  & 18.8  \\
    \rowcolor[rgb]{ .949,  .949,  .949} w/ MP & \textbf{24.3} & \textbf{49.6} & \textbf{21.9} & \textbf{8.2} & \textbf{62.5} & \textbf{58.1} & \textbf{41.9} & \textbf{28.3} \\
    \end{tabular}%
  } \vspace{-3mm}
  \caption{
  \textbf{Improvements to the baselines introduced by the \MP paradigm on the UT-Zappos dataset.}
  }
  \label{tab:baseline-MP}%
  \vspace{-5mm}
\end{table}%

\paragraph{CZSL Task Formulation.} 
Given the 
state set $\mathcal{S} = \{ s_1, s_2, \dots, s_{|\mathcal{S}|} \}$ and object set $\mathcal{O} = \{ o_1, o_2, \dots, o_{|\mathcal{O}|} \}$ as the primitive concepts, where $|\cdot|$ denotes the number of elements in the set, the compositional label space $\mathcal{C}$ is defined as
their Cartesian product,
\textit{i.e.}, $\mathcal{C} = \mathcal{S} \times \mathcal{O}$.
And the set of the seen and unseen compositions, denoted as $\mathcal{C}^{se}$ and $\mathcal{C}^{us}$, are the two disjoint subsets of $\mathcal{C}$, \textit{i.e.}, $\mathcal{C}^{se} \cap \mathcal{C}^{us} = \emptyset$.
To learn a model %
that assigns compositional labels from the target set $\mathcal{C}^{tgt}$ to the input images, a training set $\mathcal{T} = \{ (x_i, c_i) | x \in \mathcal{X}, c \in \mathcal{C}^{se} \}$ is provided, where $\mathcal{X}$ denotes the image space.    %
In the closed-world setting, the target set is defined as $\mathcal{C}^{tgt} = \mathcal{C}^{se} \cup \mathcal{C}^{us}$, where only the known composition space is considered.
And in the open-world setting, 
the target set
is all possible permutations of the state-object compositions, \textit{i.e.}, $\mathcal{C}^{tgt} = \mathcal{C}$. %

\vspace{-4mm}
\paragraph{Visual Feature Extraction.} 
Given the input image $x \in \mathbb{R}^{H \times W \times C}$, 
the image encoder $E_v$, implemented with ViT~\cite{Dosovitskiy:ViT}, first splits it into $N^p = HW/P^2$ non-overlapping patches, where $(P,P)$ is the resolution of each patch.
The patches are projected to form a sequence of patch tokens together with a pre-trained \texttt{[CLS]} token, where the pre-trained position embeddings are also added to preserve positional information.
Then, the encoder $E_v$ updates the token sequence $\mathbf{X} \in \mathbb{R}^{(N^p+1) \times d^{in}_v}$ with self-attention-based blocks, where $d^{in}_v$ is the dimension of each visual token.
Finally,
a single linear layer $g^{proj}$ with parameters $\mathbf{W}_g \in \mathbb{R}^{d^{in}_v \times d}$ projects the output \texttt{[CLS]} token, where $d$ is the dimension of the cross-modal latent space.
And the projected token $\mathbf{x}^{\texttt{CLS}} \in \mathbb{R}^{d}$ serves as the image representation.

\subsection{A Revisit of Existing Framework} \label{sec:framework-revisit}

Taking a closer look into existing CLIP-based works~\cite{Nayak:CSP,Xu:PromptCompVL,Lu:DFSP}, we found that all of them build a single cross-modal alignment for inference, which yields the recognition probability $p(c_{i,j}|x)$ given the input image $x$ and the candidate pair $c_{i,j}=\langle s_i, o_j \rangle$. Since the frozen CLIP backbone has provided a well-established vision-language alignment, 
an essential step for these methods is to construct appropriate prompts for compositional labels.
Commonly,
initializing with the pre-trained embeddings from CLIP,
a new primitive \textit{vocabulary} $\mathbf{V} = [\mathbf{V}^{\mathcal{S}}, \mathbf{V}^{\mathcal{O}}] \in \mathbb{R}^{(|\mathcal{S}|+|\mathcal{O}|) \times d^{in}_t}$ is first built for all states and objects, where $d^{in}_t$ is the dimension of each vocabulary token.
Then,
a natural language \textit{prefix} such as ``\textit{a photo of}'' is transformed into tokens with the pre-trained embeddings.
Different from the prompt format adopted in the inference of CLIP, the CZSL methods~\cite{Nayak:CSP} append the prefix tokens to the state-object composition instead of the
class
placeholder, acquiring the prompt $\mathbf{P}_{i,j} = [\mathbf{p}_1, \dots, \mathbf{p}_m, \mathbf{v}^s_i, \mathbf{v}^o_j ]$ for the pair $c_{i,j}$, where $\{\mathbf{p}_1, \dots, \mathbf{p}_m\} \in \mathbb{R}^{m \times d^{in}_t}$ are the prefix tokens, and $\mathbf{v}^s_i, \mathbf{v}^o_j$ are the vocabulary tokens of $s_i$ and $o_j$.
By feeding $\mathbf{P}_{i,j}$ into the text encoder $E_t$, the prompt representation $\mathbf{t}_{i,j} \in \mathbb{R}^{d}$ is acquired to compute $p(c_{i,j}|x)$ by cosine similarity with image representation $\mathbf{x}^{\texttt{CLS}}$.
While earlier works~\cite{Nayak:CSP,Xu:PromptCompVL} simply turn the primitive vocabulary or prefix tokens from fixed to trainable, 
DFSP~\cite{Lu:DFSP} further decomposes the prompt representations into state and object ones to provide more supervision for training. 
However, the inference still relies on a single path for estimating the composition probability.

\begin{figure*}[!t]
  \centering
   \includegraphics[width=\linewidth]{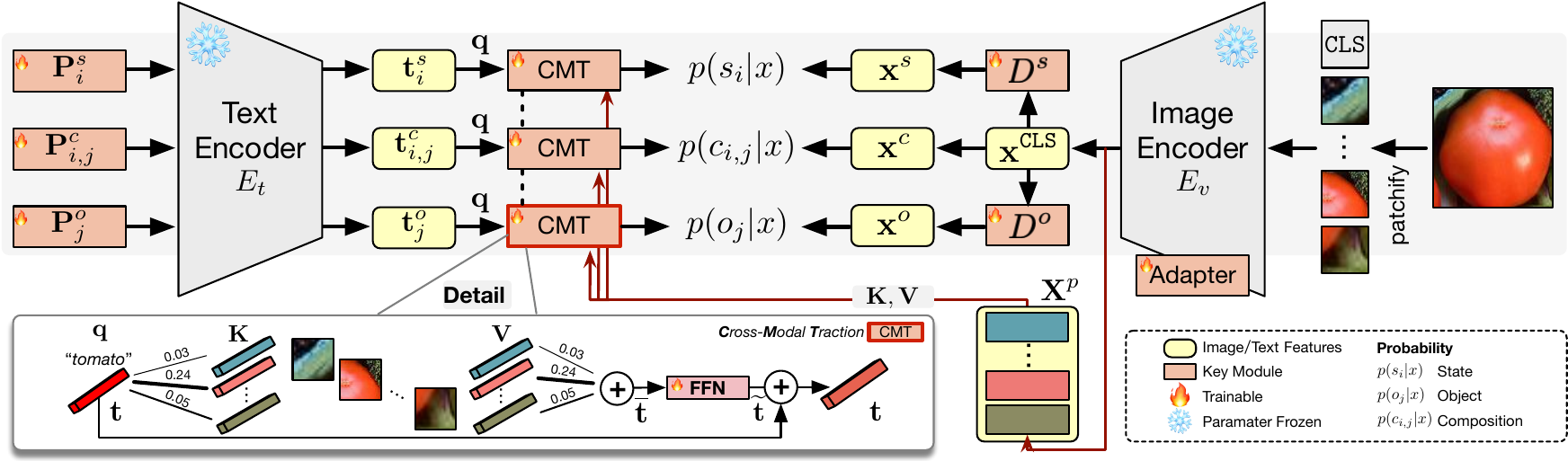}
   \vspace{-5mm}
   \caption{
   \textbf{Overview of the proposed \method.}
   }
   \label{fig:method}
   \vspace{-4mm}
\end{figure*}

\subsection{Generalizing in Multi-Path Paradigm} \label{sec:multi-path-paradigm}

As already discussed in the introduction, 
existing methods still suffer from limitations in knowledge transfer and generalization.
We believe that the issue stems from the applied single-path paradigm, and thus present a novel \MP paradigm.
An intuitive comparison of the existing and proposed paradigms is illustrated in \cref{fig:multi-path}.
Crucially, the \MP paradigm requires a recognition branch for each of the three semantic components, \textit{i.e.}, state, object, and composition.
These branches are essentially cross-modal alignments that independently unearth specific knowledge from large-scale vision-language pre-training.

Specifically, during training, three branches can collectively optimize the parameters in a multi-task learning~\cite{Zhang:multi-task-learning} manner.
And for inference, the prediction results of state and object can be incorporated to assist the composition branch. %
Formally, the integrated composition probability $\widetilde{p}(c_{i,j}|x)$ is defined as
{\setlength\abovedisplayskip{2mm}
\setlength\belowdisplayskip{2mm}
\begin{align} \label{eq:inference}
\widetilde{p}(c_{i,j}|x) = p(c_{i,j}|x) + p(s_i|x) \cdot p(o_j|x),
\end{align}}%
where the joint distribution of composition probabilities, when attribute and object predictions are considered independent of each other, is treated as a bias correction for the direct prediction of compositions.
And the most likely composition can be predicted as
{\setlength\abovedisplayskip{2mm}
\setlength\belowdisplayskip{2mm}
\begin{align} \label{eq:predict}
\hat{c} = \mathop{\arg \max}\limits_{c_{i,j} \in \mathcal{C}^{tgt}} \left(\widetilde{p}(c_{i,j}|x)\right),
\end{align}}%
thereby mitigating the excessive bias towards seen compositions and promoting a more robust recognition system.
Without further implementation constraints,
the flexibility of the \MP paradigm allows for the free derivation of new methods based on powerful VLMs.
To demonstrate the usability and effectiveness of the proposed paradigm, 
we follow it to rapidly extend the two popular baseline models,
CLIP~\cite{Radford:CLIP} and CoOp~\cite{Zhou:CoOp},
and illustrate the significant improvements in \cref{tab:baseline-MP}.

\section{Troika: An Efficient Implementation}

To adhere to the paradigm for more efficient solutions, the multi-modal feature extraction across branches must be carefully designed.
In this section, we first detail how the proposed \method develops a range of instantiations.
Then, we introduce the plug-and-play \CMT module for \method.
An overview of \method is illustrated in \cref{fig:method}.

\subsection{Instantiations}

\paragraph{Learning Prompt Representations.} 
Since the composition prompt can only elicit information related to seen compositions from VLMs, \method respectively conducts prompts for the state, object, and composition branch.
Compared to the DFSP~\cite{Lu:DFSP} approach that decomposes the text features into state and object ones, 
individual prompts for different semantic components activate the backbone from input, thus maximizing the exploitation of pre-trained knowledge.
Moreover, as the semantic roles of the target classes on each branch are different, it is natural to introduce different priors through special contexts.
Therefore, while maintaining the same primitive vocabulary as a cue of semantic compositionality, we employ an independent prompt prefix for each branch of \method.
For each state-object pair $c_{i,j}=\langle s_i, o_j \rangle$, the state prompt $\mathbf{P}^s_i$, object prompt $\mathbf{P}^o_j$, and composition prompt $\mathbf{P}^c_{i,j}$ can be constructed as
{\setlength\abovedisplayskip{1mm}
\setlength\belowdisplayskip{2mm}
\begin{align}
\mathbf{P}^s_i &= [ \mathbf{p}^s_1, \dots, \mathbf{p}^s_m, \mathbf{v}^s_i ], \\
\mathbf{P}^o_j &= [ \mathbf{p}^o_1, \dots, \mathbf{p}^o_m, \mathbf{v}^o_j ], \\
\mathbf{P}^c_{i,j} &= [ \mathbf{p}^c_1, \dots, \mathbf{p}^c_m, \mathbf{v}^s_i, \mathbf{v}^o_j ],
\end{align}}%
where $\{\mathbf{p}^s_1, \dots, \mathbf{p}^s_m\}$, $\{\mathbf{p}^o_1, \dots, \mathbf{p}^o_m\}$, and $\{\mathbf{p}^c_1, \dots, \mathbf{p}^c_m\}$ are the learnable state prefix, object prefix, and composition prefix, respectively.
These fully trainable prompts are then fed into the text encoder $E_t$ to obtain the prompt representation for each branch, formulated as
{
\setlength\abovedisplayskip{2mm}
\setlength\belowdisplayskip{2mm}
\begin{align}
\mathbf{t}^s_i = E_t(\mathbf{P}^s_i),\quad \mathbf{t}^o_j = E_t(\mathbf{P}^o_j),\quad \mathbf{t}^c_{i,j} = E_t(\mathbf{P}^c_{i,j}).
\end{align}}

\vspace{-5mm}
\paragraph{Learning Visual Representations.} 
While existing methods~\cite{Nayak:CSP,Xu:PromptCompVL,Lu:DFSP} directly apply the frozen image encoder, 
we first trial several easy-to-implement and effective PETL techniques.
Based on the experimental results, we finally introduce Adapter~\cite{Houlsby:Adapter} to adapt the image encoder without updating its original parameters.
These small neural modules (adapters)
can also be freely replaced by more complicated techniques to pursue further enhancements.
Then, to establish cross-modal alignment on each branch, specific visual features for the composition, state, and object should be extracted.
Still treating the image representation $\mathbf{x}^{\texttt{CLS}}$ as the composition visual representation $\mathbf{x}^c$, we introduce the state disentangler $D^s$ and object disentangler $D^o$ to decompose the state and object visual features $\mathbf{x}^s$ and $\mathbf{x}^o$
as
{
\setlength\abovedisplayskip{2mm}
\setlength\belowdisplayskip{2mm}
\begin{align}
\mathbf{x}^s = D^s(\mathbf{x}^{\texttt{CLS}}),\quad \mathbf{x}^o = D^o(\mathbf{x}^{\texttt{CLS}}),\quad \mathbf{x}^c = \mathbf{x}^{\texttt{CLS}},
\end{align}}%
where $D^s$ and $D^o$ are implemented with two individual MLPs.
This simple structure provides the necessary non-linear mapping to resolve primitive-specific features from entangled global features.

\vspace{-5mm}
\paragraph{Training.} 
Given the prompt representations and visual features specified with different branches, the probability of assigning labels of the state $s_i$, object $o_j$, and composition $c_{i,j}$ to the image can be computed separately as

{\setlength\abovedisplayskip{-4mm}
\setlength\belowdisplayskip{2mm}
\begin{align} 
p(s_i|x) &= \frac{\exp(\mathbf{x}^s \cdot \mathbf{t}^s_i / \tau)}{\sum^{|\mathcal{S}|}_{k=1}\exp(\mathbf{x}^s \cdot \mathbf{t}^s_k / \tau)}, \\
p(o_j|x) &= \frac{\exp(\mathbf{x}^o \cdot \mathbf{t}^o_j / \tau)}{\sum^{|\mathcal{O}|}_{k=1}\exp(\mathbf{x}^o \cdot \mathbf{t}^o_k / \tau)}, \\
p(c_{i,j}|x) &= \frac{\exp(\mathbf{x}^c \cdot \mathbf{t}^c_{i,j} / \tau)}{\sum^{|\mathcal{C}^{tgt}|}_{k=1}\exp(\mathbf{x}^c \cdot \mathbf{t}^c_k / \tau)},
\end{align}}%
where $\tau \in \mathbb{R}$ is the pre-trained temperature parameter from CLIP.
In each branch, the cross-entropy loss encourages the model to explicitly recognize the corresponding semantic role, described as
{\setlength\abovedisplayskip{1mm}
\setlength\belowdisplayskip{2mm}
\begin{align}
\mathcal{L}^s = - \frac{1}{|\mathcal{X}|} \sum_{x \in \mathcal{X}} \log p(s|x),
\end{align}}
{
\setlength\abovedisplayskip{-2mm}
\setlength\belowdisplayskip{0mm}
\begin{align}
\mathcal{L}^o = - \frac{1}{|\mathcal{X}|} \sum_{x \in \mathcal{X}} \log p(o|x),
\end{align}}
{
\setlength\abovedisplayskip{-2mm}
\setlength\belowdisplayskip{2mm}
\begin{align}
\mathcal{L}^c = - \frac{1}{|\mathcal{X}|} \sum_{x \in \mathcal{X}} \log p(c|x).
\end{align}}%
Therefore, the overall training loss $\mathcal{L}_{all}$ is defined as
{\setlength\abovedisplayskip{2mm}
\setlength\belowdisplayskip{2mm}
\begin{align} \label{eq:overall_loss}
\mathcal{L}_{all} = \alpha^s \mathcal{L}^s + \alpha^o \mathcal{L}^o + \alpha^c \mathcal{L}^c,
\end{align}}%
where $\alpha^s, \alpha^o, \alpha^c \in \mathbb{R}$ are weighting coefficients to balance the influences of different losses.
Note that we have omitted the weight decay here for simplicity.
\method adheres to the unified inference of the \MP paradigm (\textit{i.e.}, \cref{eq:inference,eq:predict}) without making any additional modifications.

\subsection{Cross-Modal Traction} \label{sec:cross-modal-traction}

Although inheriting the rich cross-modal understanding of VLMs, discrepancies may still exist between semantically similar vision-language representations.
Given the same semantic concept, the static and monotonous prompt representation naturally fails to be commonly optimal for all input images that 
come
from a plentiful distribution.
This issue becomes more serious in the additional state and object branches, as the visual content of the same primitive changes considerably when paired with different primitives. %
Therefore,
we further develop a \CMT module for \method.
The module adaptively shifts the prompt representation to accommodate the content diversity and diminish the cross-modal discrepancies.
In this process, relevant patch features serve as the guidance to avoid noise from semantic-agnostic sub-regions interfering with the traction.

Specifically, the \CMT module is composed of a stack of $N$ blocks, and in each block, we first consider a scaled dot product attention mechanism~\cite{Vaswani:Transformer} with the prompt representation attending to all patch tokens.
Given the input prompt representation $\mathbf{t}$ that comes from an arbitrary branch, we first acquire the patch tokens $\mathbf{X}^p \in \mathbb{R}^{N^p \times d}$ after projecting them with the linear layer $g^{proj}$.
Then, the query, key and value can be derived as
{\setlength\abovedisplayskip{2mm}
\setlength\belowdisplayskip{2mm}
\begin{align}
\mathbf{q} = \mathbf{t} \mathbf{W}_q,\quad \mathbf{K} = \mathbf{X}^p \mathbf{W}_K,\quad \mathbf{V} = \mathbf{X}^p \mathbf{W}_V,
\end{align}}%
where $\mathbf{W}_q, \mathbf{W}_K, \mathbf{W}_V \in \mathbb{R}^{d \times d^{attn}}$ are the parameter matrices, and $d^{attn}$ is the dimension of the single-head attention.
The dot product attention gives relevance weights from $\mathbf{t}$ to each patch token, which are used to aggregate the value-projected patch tokens as 
{\setlength\abovedisplayskip{2mm}
\setlength\belowdisplayskip{2mm}
\begin{align}
\overline{\mathbf{t}} = \text{Attention}(\mathbf{q}, \mathbf{K}, \mathbf{V}) = \text{softmax}\left( \frac{\mathbf{q}\mathbf{K}^{\top}}{\sqrt{d^{attn}}}\right)\mathbf{V}.
\end{align}}%
In practice, a multi-head design with $h=d/d^{attn}$ parallel attention heads is naturally introduced to diversify representation subspaces.
After the attention layer, a feed-forward network $\text{FFN}$, implemented as a MLP, is introduced as
{\setlength\abovedisplayskip{2mm}
\setlength\belowdisplayskip{2mm}
\begin{align}
\widetilde{\mathbf{t}} = \text{FFN}(\overline{\mathbf{t}}) = \sigma(\overline{\mathbf{t}}\mathbf{W}_1 + \mathbf{b}_1)\mathbf{W}_2 + \mathbf{b}_2,
\end{align}}%
where $\mathbf{W}_1, \mathbf{W}_2$ are parameter matrices, $\mathbf{b}_1, \mathbf{b}_2$ are bias terms, and $\sigma(\cdot)$ is a nonlinear activation function.
Note that for both the attention and the feed-forward network, we omit the residual connections around them for simplicity.
Then, we can update the prompt representation as
{\setlength\abovedisplayskip{2mm}
\setlength\belowdisplayskip{2mm}
\begin{align}
\mathbf{t} \leftarrow \mathbf{t} + \boldsymbol{\lambda} \cdot \widetilde{\mathbf{t}},
\end{align}}%
where $\boldsymbol{\lambda} \in \mathbb{R}^d$ is a trainable parameter vector controlling the strength of the cross-modal traction in each dimension.
$\widetilde{\mathbf{t}}$ can be viewed as a traction that pulls $\mathbf{t}$ towards the visual content.
And the whole module can be seamlessly inserted into each branch before calculating the cross-modal matching probability.
In practice, all three branches share the same module to reduce the parameter overhead.

\section{Experiments}

\subsection{Experimental Setup}

\paragraph{Datasets.}
We experiment with three real-world CZSL benchmarks: MIT-States~\cite{Isola:MIT-States}, UT-Zappos~\cite{Yu:UT-Zappos}, and C-GQA~\cite{Naeem:CGE}.
We follow the split suggested by previous works~\cite{Purushwalkam:task-driven-modular-networks,Nayak:CSP}, and summarize detailed statistics in~\cref{tab:dataset_stats}.

\begin{table}[!t]
\tablestyle{5pt}{1.0}
\setlength\tabcolsep{1.9pt}
\def\w{20pt} 
\scalebox{1}{
    \begin{tabular}{lcc|cc|ccc|ccc}
          &       &       & \multicolumn{2}{c|}{\textbf{Training}} & \multicolumn{3}{c|}{\textbf{Validation}} & \multicolumn{3}{c}{\textbf{Test}} \\
    \textbf{Dataset} & $\mathcal{S}$ & $\mathcal{O}$ & $\mathcal{C}^{se}$ & $\mathcal{X}$ & $\mathcal{C}^{se}$ & $\mathcal{C}^{us}$ & $\mathcal{X}$ & $\mathcal{C}^{se}$ & $\mathcal{C}^{us}$ & $\mathcal{X}$ \\
    \shline
    MIT-States~\cite{Isola:MIT-States} & 115   & 245   & 1262  & 30k   & 300   & 300   & 10k   & 400   & 400   & 13k \\
    UT-Zappos~\cite{Yu:UT-Zappos} & 16    & 12    & 83    & 23k   & 15    & 15    & 3k    & 18    & 18    & 3k \\
    C-GQA~\cite{Naeem:CGE} & 413     & 674     & 5592    & 27k    & 1252    & 1040    & 7k    & 888     & 923     & 5k \\
    \end{tabular}%
    } \vspace{-2mm}
  \caption{
  \textbf{Statistics of three datasets in our experiments.} 
  The number of elements in each set is reported.
  }
  \label{tab:dataset_stats}%
  \vspace{-5mm}
\end{table}%

\begin{table*}[!t]    %
\tablestyle{5pt}{1.0}
\setlength\tabcolsep{1pt}
\def\w{20pt} 
\scalebox{1}{
    \begin{tabular}{lcccc|cccc|cccc}
    \multirow{2}[0]{*}{\textbf{Method}} & \multicolumn{4}{c|}{\textbf{MIT-States}} & \multicolumn{4}{c|}{\textbf{UT-Zappos}} & \multicolumn{4}{c}{\textbf{C-GQA}} \\[-1.5pt]
          & S     & U     & HM     & AUC   & S     & U     & HM     & AUC   & S     & U     & HM     & AUC \\
    \shline
    \multicolumn{13}{c}{ \demph{ \it{\textbf{Closed-world} Results} } }\\
    \hline
    CLIP~\cite{Radford:CLIP}  & 30.2  & 46.0  & 26.1  & 11.0  & 15.8  & 49.1  & 15.6  & 5.0   & 7.5   & 25.0  & 8.6   & 1.4 \\
    CoOp~\cite{Zhou:CoOp}  & 34.4  & 47.6  & 29.8  & 13.5  & 52.1  & 49.3  & 34.6  & 18.8  & 20.5  & 26.8  & 17.1  & 4.4 \\
    CSP~\cite{Nayak:CSP}   & 46.6  & 49.9  & 36.3  & 19.4  & 64.2  & 66.2  & 46.6  & 33.0  & 28.8  & 26.8  & 20.5  & 6.2 \\
    PromptCompVL~\cite{Xu:PromptCompVL} & 48.5  & 47.2  & 35.3  & 18.3  & 64.4  & 64.0  & 46.1  & 32.2  & - & - & - & - \\
    DFSP(i2t)~\cite{Lu:DFSP} & 47.4  & 52.4  & 37.2  & 20.7  & 64.2  & 66.4  & 45.1  & 32.1  & 35.6  & 29.3  & 24.3  & 8.7 \\
    DFSP(BiF)~\cite{Lu:DFSP} & 47.1  & 52.8  & 37.7  & 20.8  & 63.3  & 69.2  & 47.1  & 33.5  & 36.5  & 32.0  & 26.2  & 9.9 \\
    DFSP(t2i)~\cite{Lu:DFSP} & 46.9  & 52.0  & 37.3  & 20.6  & 66.7  & 71.7  & 47.2  & 36.0  & 38.2  & 32.0  & 27.1  & 10.5 \\
    \rowcolor[rgb]{ .949,  .949,  .949} \textbf{\method (Ours)} & \textbf{49.0}\tiny{$\pm$0.4} & \textbf{53.0}\tiny{$\pm$0.2} & \textbf{39.3}\tiny{$\pm$0.2} & \textbf{22.1}\tiny{$\pm$0.1} & \textbf{66.8}\tiny{$\pm$1.1} & \textbf{73.8}\tiny{$\pm$0.6} & \textbf{54.6}\tiny{$\pm$0.5} & \textbf{41.7}\tiny{$\pm$0.7} & \textbf{41.0}\tiny{$\pm$0.2} & \textbf{35.7}\tiny{$\pm$0.3} & \textbf{29.4}\tiny{$\pm$0.2} & \textbf{12.4}\tiny{$\pm$0.1} \\
    \hline\\[-2.4ex]
    \multicolumn{13}{c}{ \demph{ \it{\textbf{Open-world} Results} } }\\
    \hline
    CLIP~\cite{Radford:CLIP}  & 30.1  & 14.3  & 12.8  & 3.0   & 15.7  & 20.6  & 11.2  & 2.2   & 7.5   & 4.6   & 4.0   & 0.27 \\
    CoOp~\cite{Zhou:CoOp}  & 34.6  & 9.3   & 12.3  & 2.8   & 52.1  & 31.5  & 28.9  & 13.2  & 21.0  & 4.6   & 5.5   & 0.70 \\
    CSP~\cite{Nayak:CSP}   & 46.3  & 15.7  & 17.4  & 5.7   & 64.1  & 44.1  & 38.9  & 22.7  & 28.7  & 5.2   & 6.9   & 1.20 \\
    PromptCompVL~\cite{Xu:PromptCompVL} & 48.5  & 16.0  & 17.7  & 6.1   & 64.6  & 44.0  & 37.1  & 21.6  & - & - & - & - \\
    DFSP(i2t)~\cite{Lu:DFSP} & 47.2  & 18.2  & 19.1  & 6.7   & 64.3  & 53.8  & 41.2  & 26.4  & 35.6  & 6.5   & 9.0   & 1.95 \\
    DFSP(BiF)~\cite{Lu:DFSP} & 47.1  & 18.1  & 19.2  & 6.7   & 63.5  & 57.2  & 42.7  & 27.6  & 36.4  & 7.6   & 10.6  & 2.39 \\
    DFSP(t2i)~\cite{Lu:DFSP} & 47.5  & 18.5  & 19.3  & 6.8   & \textbf{66.8} & 60.0  & 44.0  & 30.3  & 38.3  & 7.2   & 10.4  & 2.40 \\
    \rowcolor[rgb]{ .949,  .949,  .949} \textbf{\method (Ours)} & \textbf{48.8}\tiny{$\pm$0.4} & \textbf{18.7}\tiny{$\pm$0.1} & \textbf{20.1}\tiny{$\pm$0.1} & \textbf{7.2}\tiny{$\pm$0.1} & 66.4\tiny{$\pm$1.0}  & \textbf{61.2}\tiny{$\pm$1.0} & \textbf{47.8}\tiny{$\pm$1.3} & \textbf{33.0}\tiny{$\pm$1.0} & \textbf{40.8}\tiny{$\pm$0.2} & \textbf{7.9}\tiny{$\pm$0.2} & \textbf{10.9}\tiny{$\pm$0.3} & \textbf{2.70}\tiny{$\pm$0.1} \\
    \end{tabular}%
  }\vspace{-2mm}
  \caption{
  \textbf{Main results on three benchmarks.} 
  All methods use a CLIP ViT-L/14 backbone.
  For our \method, we report the average performance on 5 random seeds with standard error.}
  \label{tab:main-SOTA}%
  \vspace{-5mm}
\end{table*}%

\vspace{-5mm}
\paragraph{Metrics.}
We follow the evaluation protocol of previous works~\cite{Purushwalkam:task-driven-modular-networks,Naeem:CGE,Tian:IVR}, where a calibration bias trades off between the prediction scores of seen and unseen pairs at test time.
While varying the candidate bias from $-\infty$ to $+\infty$, a curve can be drawn with the accuracy of seen and unseen pairs.
To quantify the overall performance on both seen and unseen pairs, we compute the area under the curve (\textbf{AUC}) and find the point with the best harmonic mean (\textbf{HM}) between the seen and unseen accuracy.
We also report the best seen accuracy (\textbf{S}) by adjusting the bias to $-\infty$, and the best unseen accuracy (\textbf{U}) by adjusting the bias to $+\infty$.

\vspace{-5mm}
\paragraph{Implementation Details.}
We implement \method with a pre-trained CLIP ViT-L/14 model in PyTorch~\cite{Paszke:PyTorch}.
The model is trained and evaluated on an NVIDIA A100 GPU.
For the open-world evaluation, we follow the post-training calibration method~\cite{Nayak:CSP} to filter out infeasible compositions.
More details can be found in the supplementary material.

\subsection{Main Results}

For a fair comparison, we primarily compare with CLIP-based methods using the same CLIP ViT-L/14 backbone.
In particular, pre-trained CLIP~\cite{Radford:CLIP}, CoOp~\cite{Zhou:CoOp}, CSP~\cite{Nayak:CSP}, PromptCompVL~\cite{Xu:PromptCompVL}, and all versions of DFSP~\cite{Lu:DFSP} are considered.
For comparisons with more baselines involved, please refer to 
the supplementary material.

In \cref{tab:main-SOTA}, we report both closed-world and open-world results.
On the \textbf{closed-world} setting, our \method exceeds the previous SOTA methods on MIT-States, UT-Zappos, and C-GQA.
Specifically, relative to existing methods, \method improves the HM by +2.0\%, +7.4\%, +2.3\%, and the AUC by +1.5\%, +5.7\%, +1.9\% respectively on three datasets.
And \method also achieves the best seen and unseen accuracies on these datasets.
On the \textbf{open-world} setting, our \method also achieves the SOTA results on the three datasets in terms of almost all metrics.
The only exception is that the best seen accuracy of \method is 0.4\% lower than DFSP(t2i)~\cite{Lu:DFSP}.
However, \method outperforms the existing methods by +0.8\%, +3.8\%, +0.5\% in terms of the HM, and by +0.4\%, +2.7\%, +0.3\% in terms of the AUC on the three datasets, indicating that our \method achieves a more consistent and comprehensive performance.

\subsection{Ablation Study} \label{sec:ablation-study}

To empirically show the effectiveness of our framework design, we conduct extensive experiments and report the closed-world results for the ablation study.

\begin{table}[!t]
\tablestyle{5pt}{1.0}
\setlength\tabcolsep{4pt}
\def\w{20pt} 
\scalebox{1}{
    \begin{tabular}{ccc|cccc|cccc}
    \multicolumn{3}{c|}{\textbf{Branch}} & \multicolumn{4}{c|}{\textbf{UT-Zappos}} & \multicolumn{4}{c}{\textbf{C-GQA}} \\
    $c$  & $s$ & $o$ & S     & U     & HM    & AUC   & S     & U     & HM    & AUC \\
    \shline
    \multicolumn{11}{c}{ \demph{ \it{Training + Inference} } }\\
    \hline
    \rowcolor[rgb]{ .949,  .949,  .949} \cmark & \cmark & \cmark & 66.8  & 73.8  & \textbf{54.6} & \textbf{41.7} & \textbf{41.0} & \textbf{35.7} & \textbf{29.4} & \textbf{12.4} \\    
    \cmark &       &       & 66.8  & \textbf{74.5}  & 49.9  & 37.7  & 39.6  & 34.3  & 28.9  & 11.6 \\
          & \cmark & \cmark & 67.9  & 69.4  & 47.0  & 35.7  & 35.8  & 20.2  & 19.1  & 5.9 \\
    \cmark & \cmark &       & 63.4  & 73.9 & 51.1  & 38.6  & 40.5  & 34.4  & 28.8  & 11.8 \\
    \cmark &       & \cmark & 67.2  & 73.1  & 49.7  & 37.7  & 39.5  & 33.8  & 28.8  & 11.5 \\
    \hline
    \multicolumn{11}{c}{ \demph{ \it{Inference} } }\\
    \hline
    \rowcolor[rgb]{ .949,  .949,  .949} \cmark & \cmark & \cmark & 66.8  & 73.8  & \textbf{54.6} & \textbf{41.7} & \textbf{41.0} & \textbf{35.7} & \textbf{29.4} & \textbf{12.4} \\        
    \cmark &       &       & \textbf{68.2} & 72.1  & 53.0  & 41.0  & 39.6  & 34.0  & 28.3  & 11.5 \\
          & \cmark & \cmark & 66.4  & 68.8  & 46.5  & 34.5  & 36.9  & 20.7  & 19.8  & 6.3 \\
    \cmark & \cmark &       & 66.3  & 70.0  & 53.7  & 39.9  & 40.7  & 33.7  & 28.8  & 11.7 \\
    \cmark &       & \cmark & \textbf{68.2} & 72.9  & 52.0  & 40.4  & 40.0  & 34.4  & 28.4  & 11.7 \\
    \end{tabular}%
  } \vspace{-2mm}
  \caption{\textbf{Ablation on the \MP paradigm.}
  The best results are obtained by keeping all three branches in both the training and inference phases.
  }
  \label{tab:ablation-branch}%
  \vspace{-7mm}
\end{table}%

\vspace{-5mm}
\paragraph{Ablation on Multi-Path Paradigm.}
In \cref{tab:ablation-branch}, we remove one or more specific branches at a time to prove that all branches in the \MP paradigm contribute.
Specifically, two scenarios are considered:
(1) \textbf{Training + Inference}, which refers to simultaneously eliminating the effects of the corresponding branches in both training and inference phases, \textit{i.e.}, removing the corresponding losses from \cref{eq:overall_loss} and the corresponding probabilities from \cref{eq:inference}.
(2) \textbf{Inference}, which refers to eliminating the effects of the corresponding branches only during inference, leaving the training loss unchanged.
Several observations in the table are worth highlighting:
(1) In both scenarios, removing the composition branch results in the greatest drop in performance, illustrating the importance of the branch for learning compositionality.
(2) Generally, removing the branches only during inference achieves a better result than removing them in both phases, indicating that all loss items have a positive impact.  %
(3) Keeping all branches in both scenarios leads to the best HM and AUC.
Note that some cases on UT-Zappos achieve a higher best seen or unseen accuracy, which only means that they might be better in unrealistic extremes.
And their worse HM and AUC suggest that removing branches would introduce instability.

\begin{table}[!t]
\tablestyle{5pt}{1.0}
\setlength\tabcolsep{3pt}
\def\w{20pt} 
\scalebox{1}{
    \begin{tabular}{cc|cccc|cccc}
          &       & \multicolumn{4}{c|}{\textbf{UT-Zappos}} & \multicolumn{4}{c}{\textbf{C-GQA}} \\
    \textbf{Prefix} & \textbf{Vocab.} & S     & U     & HM    & AUC   & S     & U     & HM    & AUC \\
    \shline
    \rowcolor[rgb]{ .949,  .949,  .949} $c|s|o$ & $cso$   & \textbf{66.8} & \textbf{73.8} & \textbf{54.6} & \textbf{41.7} & \textbf{41.0} & \textbf{35.7} & \textbf{29.4} & \textbf{12.4} \\
    $cso$   & $cso$   & \textbf{66.8} & 72.8  & 54.0  & 41.1  & 39.6  & 32.9  & 28.7  & 11.3 \\
    $c|so$  & $cso$   & 66.2  & 72.9  & 54.5  & 41.4  & 39.8  & 33.7  & 28.9  & 11.6 \\
    $c|s|o$ & $c|s|o$ & 67.2  & 72.5  & 52.9  & 40.8  & 39.9  & 33.0  & 28.9  & 11.5 \\
    \end{tabular}%
  } \vspace{-3mm}
  \caption{\textbf{Ablation on individual prefixes and shared vocabulary.}
  Branches separated by ``$|$'' do not share the corresponding prompt parameters.
  }
  \label{tab:ablation-prompt}%
  \vspace{-2mm}
\end{table}%

\begin{table}[!t]
\tablestyle{5pt}{1.0}
\setlength\tabcolsep{4pt}
\def\w{20pt} 
\scalebox{1}{
    \begin{tabular}{c|cccc|cccc}
    \multirow{2}[1]{*}{\textbf{\method}} & \multicolumn{4}{c|}{\textbf{UT-Zappos}} & \multicolumn{4}{c}{\textbf{C-GQA}} \\
          & S     & U     & HM    & AUC   & S     & U     & HM    & AUC \\
    \shline
    \rowcolor[rgb]{ .949,  .949,  .949} w/ CMT & \textbf{66.8} & \textbf{73.8} & \textbf{54.6} & \textbf{41.7} & \textbf{41.0} & \textbf{35.7} & \textbf{29.4} & \textbf{12.4} \\
    w/o CMT & 64.4  & 70.7  & 51.9  & 37.8  & 38.5  & 33.2  & 27.9  & 11.0 \\
    \end{tabular}%
  } \vspace{-3mm}
  \caption{\textbf{Ablation on the \CMT module.} 
  }
  \label{tab:ablation-fusion}%
  \vspace{-5mm}
\end{table}%

\vspace{-5mm}
\paragraph{Ablation on Prefix and Vocabulary Design.}
In \cref{tab:ablation-prompt}, based on the current design of \method (the top row), we first allow all three branches to share the prefix parameters (the second row), as well as allowing only the state and object branches to share the prefix parameters (the third row). 
We can observe that separating the prefixes of the composition branch and the primitive branches leads to higher HM and AUC.
Moreover, maintaining an individual prefix for each branch achieves the best results for all metrics.
The results show that it is necessary to inject branch-specific prior knowledge into the prefix parameters.
We also attempt to build an individual primitive vocabulary for each branch (the bottom row), which results in a significant drop in performance.
We attribute this to a disruption of the semantic dependency modeling.
As a conclusion, jointly optimizing the shared vocabulary parameters from multiple paths contributes to the compositional learning.

\vspace{-5mm}
\paragraph{Ablation on Cross-Modal Traction Module.}
In \cref{tab:ablation-fusion}, we validate the effectiveness of our \CMT module by removing it from \method.
We can observe that equipping \method with the \CMT module boosts the HM by 2.1\% and the AUC by 2.65\% in average.
This illustrates that by effectively calibrating cross-modal deviations, the adaptive traction improves the accuracy.

\begin{figure}[!t]
  \centering
   \includegraphics[width=0.98\linewidth]{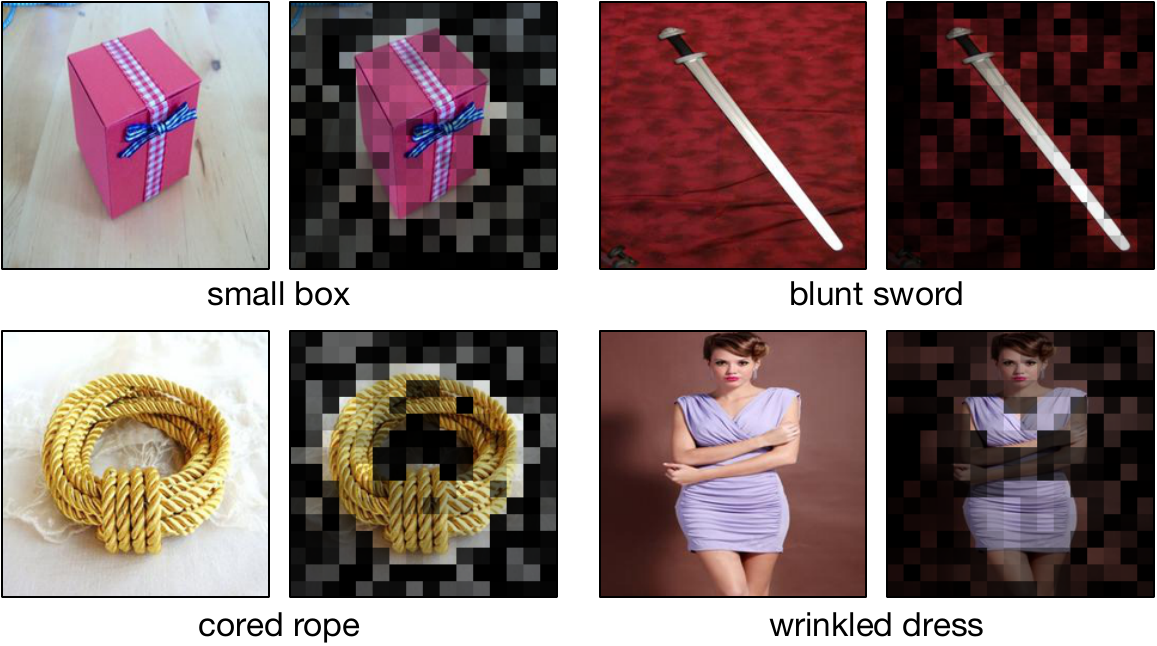}
   \vspace{-2mm}
   \caption{
   \textbf{Visualization analysis of the \CMT module.}
   We show the original image and the visualization result in pairs.
   The brighter the patch, the greater its role in the traction.
   }
   \label{fig:attn_visualization}
   \vspace{-4mm}
\end{figure}

To qualitatively evaluate whether the \CMT module indeed exploits the semantically similar patch features, we also visualize the attention weights of several test samples from MIT-States in \cref{fig:attn_visualization}.
We can observe that the patches that are closer to the label semantics receive more attention,
which also means that they contribute more to the cross-modal traction.

\begin{table*}[!t]    %
\tablestyle{5pt}{1.0}
\setlength\tabcolsep{4pt}
\def\w{20pt} 
\scalebox{1}{
    \begin{tabular}{l|cccc|cccc|cccc}
    \multirow{2}[1]{*}{\textbf{Visual Tuning}} & \multicolumn{4}{c|}{\textbf{MIT-States}} & \multicolumn{4}{c|}{\textbf{UT-Zappos}} & \multicolumn{4}{c}{\textbf{C-GQA}} \\
          & S     & U     & HM    & AUC   & S     & U     & HM    & AUC   & S     & U     & HM    & AUC \\
    \shline
    None  & 48.3  & 50.8  & 37.5  & 20.6  & 62.7  & 70.7  & 50.3  & 36.2  & 34.8  & 29.5  & 24.2  & 8.5 \\
    Full  & 41.7  & 36.3  & 28.7  & 12.2  & 48.9  & 57.4  & 34.4  & 19.1  & \textbf{44.5} & \textbf{36.5} & \textbf{31.8} & \textbf{14.1} \\
    Bias~\cite{Cai:ft-bias}  & 48.6  & \uline{52.4}  & \uline{38.8}  & \uline{21.7}  & \textbf{66.8} & 70.4  & 51.1  & 38.1  & 37.4  & 32.9  & 27.0  & 10.3 \\
    Proj~\cite{Jia:VPT}  & 47.9  & 51.6  & 38.4  & 20.9  & 63.9  & \uline{71.4}  & 52.3  & 38.9  & 35.5  & 29.1  & 24.5  & 8.7 \\
    Partial~\cite{Jia:VPT} & \textbf{49.9} & 51.3  & 38.0  & 21.4  & \uline{65.1}  & 70.8  & \uline{53.9}  & \uline{39.3}  & 38.4  & 33.3  & 28.1  & 11.1 \\
    Prompt~\cite{Jia:VPT} & 48.9  & 51.3  & 38.1  & 21.3  & 65.0  & 71.2  & 51.1  & 38.0  & 36.7  & 30.6  & 26.1  & 9.6 \\
    \rowcolor[rgb]{ .949,  .949,  .949} \textbf{Adapter}~\cite{Houlsby:Adapter} & \uline{49.0}  & \textbf{53.0} & \textbf{39.3} & \textbf{22.1} & \textbf{66.8} & \textbf{73.8} & \textbf{54.6} & \textbf{41.7} & \uline{41.0}  & \uline{35.7}  & \uline{29.4}  & \uline{12.4} \\
    \end{tabular}%
  } \vspace{-2mm}
  \caption{
  \textbf{Ablation on visual tuning strategy.}
  Best results are displayed in \textbf{boldface}, and second best results are \uline{underlined}.
  }
  \label{tab:ablation-visual-tuning}%
  \vspace{-1mm}
\end{table*}%

\begin{figure*}[!t]
  \centering
   \includegraphics[width=\linewidth]{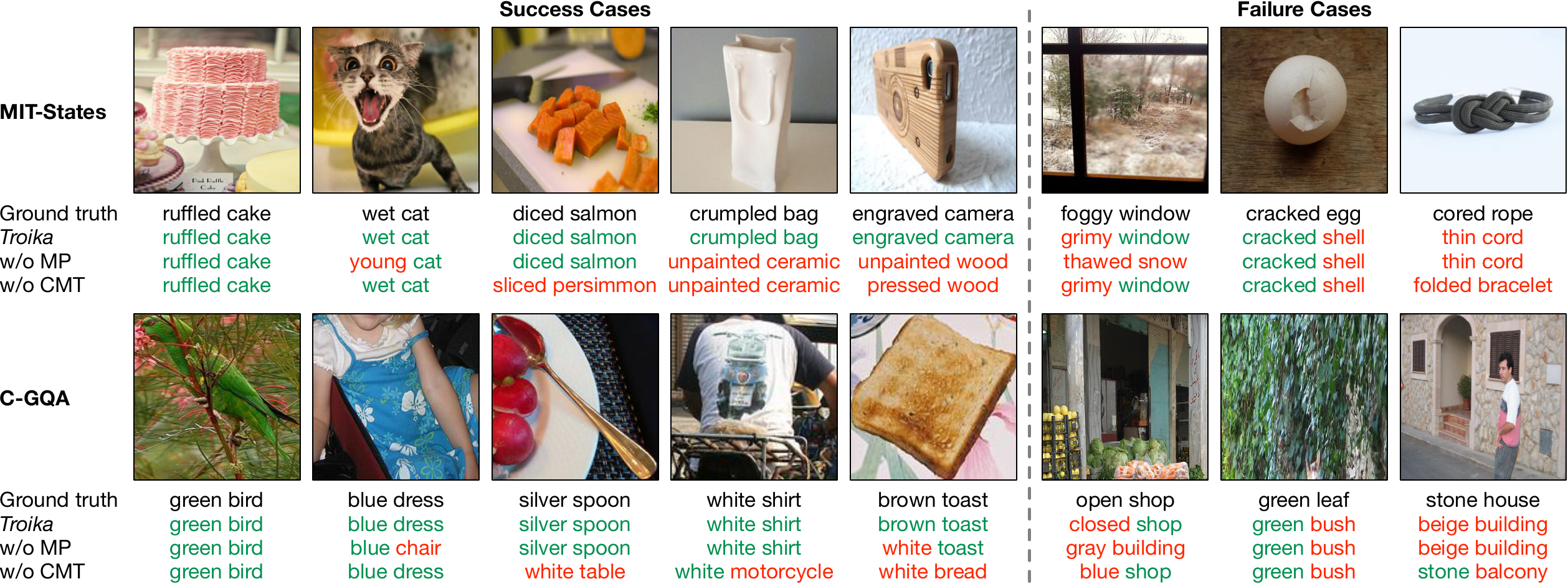}
   \vspace{-6mm}
   \caption{
   \textbf{Qualitative results.}
   We show top-1 predictions for randomly selected cases from MIT-States (the top row) and C-GQA (the bottom row).
   The complete \method correctly predicts the examples of five cols on the left, and fails on the examples of three cols on the right.
   Predictions when removing the \MP paradigm (\textit{i.e.}, w/o MP) or the \CMT module (\textit{i.e.}, w/o CMT) are also reported.
   \textcolor{rightcolor}{Green} denotes the correct prediction and \textcolor{wrongcolor}{red} denotes the wrong prediction.
   }
   \label{fig:qualitative_results}
   \vspace{-3mm}
\end{figure*}

\vspace{-5mm}
\paragraph{Ablation on Visual Tuning Strategy.} \label{sec:ablation-visual-tuning}
In \cref{tab:ablation-visual-tuning}, we compare the following popular tuning strategies for the pre-trained image encoder:
(1) \textbf{None}: freeze the encoder without updating its parameters.
(2) \textbf{Full}: fully update all parameters of the encoder.
(3) \textbf{Bias}~\cite{Cai:ft-bias,Zaken:ft-bias}: fine-tune only the bias terms.
(4) \textbf{Proj}~\cite{Jia:VPT}: fine-tune only the last linear projection layer $g^{proj}$.
(5) \textbf{Partial}~\cite{Jia:VPT}: fine-tune only the last block of the Transformer inside the encoder.
(6) \textbf{Prompt}~\cite{Jia:VPT}: fine-tune only the trainable prompt tokens inserted into the token sequence $\mathbf{X}$.
(7) \textbf{Adapter}~\cite{Houlsby:Adapter,Chen:AdaptFormer}: fine-tune only the adapter modules inserted into the Transformer inside the encoder, which is currently applied by \method.

We here highlight some observations from the table:
(1) Without adopting any tuning strategies for the image encoder, our approach still outperforms existing CLIP-based methods on most datasets, demonstrating the effectiveness of our proposed innovations including the multi-patch paradigm and \CMT mechanism.
(2) Although fully fine-tuning the image encoder achieves the best results on C-GQA, it hurts the performance on MIT-States and UT-Zappos to even underperform freezing the encoder.
Since C-GQA has much more training classes than the other two datasets, the observation suggests that fully fine-tuning the large pre-trained model easily overfits the training data, which results in poor generalization.
We note that this conclusion is consistent with existing studies~\cite{Sung:overfitting,Nayak:CSP}.
(3) All parameter-efficient tuning strategies, including those tune part of the original parameters (Bias, Proj, Partial) and tune additional parameters (Prompt, Adapter), significantly boost the performance compared to freezing the image encoder.
(4) Our applied Adapter achieves the most best results while its performance remains in the top two, indicating the superiority of our method design.

\subsection{Qualitative Results}

In \cref{fig:qualitative_results}, we visualize some qualitative results for both seen and unseen compositions, where the showed cases are randomly sampled from the test set of MIT-States and C-GQA datasets.
We report the predictions of the complete \method
and the models that remove the \MP paradigm or the \CMT module.
It can be observed that benefiting from both two innovations, \method recognizes the compositions with higher accuracy.    %
Taking the 5th case in the top row as an example, %
while the incomplete methods may be confused by the color and material presented by the object, the complete \method can focus on details such as shape, surface texture, and even local regions containing lens for comprehensive reasoning.
We also show some failure cases, %
where the entanglement of visual features places extreme demands on combinatorial understanding.
However, the proposed \MP paradigm enables the complete \method to correctly identify part of the contained primitives.
For the cases of complete prediction errors, although different from the provided labels, we find that the predictions can also interpret the content of these images.
This indicates the effectiveness of our \method from another perspective beyond the metrics.

\section{Conclusion}

In this paper, we explore the universal solution of adapting pre-trained VLMs for the downstream CZSL task.
We first propose a novel and flexible \MP paradigm that requires the simultaneous and explicit modeling of the state, object, and composition.
On top of that, follow-up researches can easily generate various new approaches with different multi-modal features.
And we develop a method named \method, which implements the paradigm with 
sophisticated
designs on both the language and vision sides.
We also present a \CMT module to improve \method by calibrating the bias between the static prompt representation and diverse visual content.
Both closed-world and open-world results on three benchmarks illustrate the superiority of \method, and extensive ablations also demonstrate the effectiveness of each component.
We hope that our work can inspire future research on exploiting foundational VLMs for compositional learning.

\noindent \textbf{Acknowledgement}
This work was supported by STI 2030—Major Projects (2022ZD0208800), NSFC General Program (Grant No. 62176215).
This work was supported by Alibaba Group through Alibaba Research Intern Program.

{
    \small
    \bibliographystyle{ieeenat_fullname}
    \bibliography{main}

\begin{thebibliography}{43}
\providecommand{\natexlab}[1]{#1}
\providecommand{\url}[1]{\texttt{#1}}
\expandafter\ifx\csname urlstyle\endcsname\relax
  \providecommand{\doi}[1]{doi: #1}\else
  \providecommand{\doi}{doi: \begingroup \urlstyle{rm}\Url}\fi

\bibitem[Anwaar et~al.(2022)Anwaar, Pan, and Kleinsteuber]{Anwaar:CVGAE}
Muhammad~Umer Anwaar, Zhihui Pan, and Martin Kleinsteuber.
\newblock On leveraging variational graph embeddings for open world compositional zero-shot learning.
\newblock In \emph{Proceedings of the ACM International Conference on Multimedia}, pages 4645--4654, 2022.

\bibitem[Cai et~al.(2020)Cai, Gan, Zhu, and Han]{Cai:ft-bias}
Han Cai, Chuang Gan, Ligeng Zhu, and Song Han.
\newblock {TinyTL:} reduce memory, not parameters for efficient on-device learning.
\newblock In \emph{Proceedings of the Advances in Neural Information Processing Systems}, pages 11285--11297, 2020.

\bibitem[Chen et~al.(2022)Chen, Ge, Tong, Wang, Song, Wang, and Luo]{Chen:AdaptFormer}
Shoufa Chen, Chongjian Ge, Zhan Tong, Jiangliu Wang, Yibing Song, Jue Wang, and Ping Luo.
\newblock {AdaptFormer:} adapting vision transformers for scalable visual recognition.
\newblock In \emph{Proceedings of the Advances in Neural Information Processing Systems}, pages 16664--16678, 2022.

\bibitem[Ding et~al.(2022)Ding, Qin, Yang, Wei, Yang, Su, Hu, Chen, Chan, Chen, Yi, Zhao, Wang, Liu, Zheng, Chen, Liu, Tang, Li, and Sun]{Ding:delta-tuning}
Ning Ding, Yujia Qin, Guang Yang, Fuchao Wei, Zonghan Yang, Yusheng Su, Shengding Hu, Yulin Chen, Chi{-}Min Chan, Weize Chen, Jing Yi, Weilin Zhao, Xiaozhi Wang, Zhiyuan Liu, Hai{-}Tao Zheng, Jianfei Chen, Yang Liu, Jie Tang, Juanzi Li, and Maosong Sun.
\newblock {Delta} {Tuning:} {A} comprehensive study of parameter efficient methods for pre-trained language models.
\newblock \emph{arXiv preprint arXiv:2203.06904}, 2022.

\bibitem[Dosovitskiy et~al.(2021)Dosovitskiy, Beyer, Kolesnikov, Weissenborn, Zhai, Unterthiner, Dehghani, Minderer, Heigold, Gelly, Uszkoreit, and Houlsby]{Dosovitskiy:ViT}
Alexey Dosovitskiy, Lucas Beyer, Alexander Kolesnikov, Dirk Weissenborn, Xiaohua Zhai, Thomas Unterthiner, Mostafa Dehghani, Matthias Minderer, Georg Heigold, Sylvain Gelly, Jakob Uszkoreit, and Neil Houlsby.
\newblock An image is worth 16x16 words: Transformers for image recognition at scale.
\newblock In \emph{Proceedings of the International Conference on Learning Representations}, 2021.

\bibitem[He et~al.(2022)He, Zhou, Ma, Berg{-}Kirkpatrick, and Neubig]{He:unify-PET}
Junxian He, Chunting Zhou, Xuezhe Ma, Taylor Berg{-}Kirkpatrick, and Graham Neubig.
\newblock Towards a unified view of parameter-efficient transfer learning.
\newblock In \emph{Proceedings of the International Conference on Learning Representations}, 2022.

\bibitem[He et~al.(2016)He, Zhang, Ren, and Sun]{He:ResNet}
Kaiming He, Xiangyu Zhang, Shaoqing Ren, and Jian Sun.
\newblock Deep residual learning for image recognition.
\newblock In \emph{Proceedings of the IEEE/CVF Conference on Computer Vision and Pattern Recognition}, pages 770--778, 2016.

\bibitem[Hendrycks and Gimpel(2016)]{Hendrycks:GELU}
Dan Hendrycks and Kevin Gimpel.
\newblock {Gaussian Error Linear Unit (GELUs)}.
\newblock \emph{arXiv preprint arXiv:1606.08415}, 2016.

\bibitem[Houlsby et~al.(2019)Houlsby, Giurgiu, Jastrzebski, Morrone, de~Laroussilhe, Gesmundo, Attariyan, and Gelly]{Houlsby:Adapter}
Neil Houlsby, Andrei Giurgiu, Stanislaw Jastrzebski, Bruna Morrone, Quentin de Laroussilhe, Andrea Gesmundo, Mona Attariyan, and Sylvain Gelly.
\newblock Parameter-efficient transfer learning for {NLP}.
\newblock In \emph{Proceedings of the International Conference on Machine Learning}, pages 2790--2799, 2019.

\bibitem[Huang et~al.(2023)Huang, Gong, Pan, Jiang, Lv, Li, and Wang]{Huang:VoP}
Siteng Huang, Biao Gong, Yulin Pan, Jianwen Jiang, Yiliang Lv, Yuyuan Li, and Donglin Wang.
\newblock {VoP:} text-video co-operative prompt tuning for cross-modal retrieval.
\newblock In \emph{Proceedings of the IEEE/CVF Conference on Computer Vision and Pattern Recognition}, pages 6565--6574, 2023.

\bibitem[Isola et~al.(2015)Isola, Lim, and Adelson]{Isola:MIT-States}
Phillip Isola, Joseph~J. Lim, and Edward~H. Adelson.
\newblock Discovering states and transformations in image collections.
\newblock In \emph{Proceedings of the IEEE/CVF Conference on Computer Vision and Pattern Recognition}, pages 1383--1391, 2015.

\bibitem[Jia et~al.(2022)Jia, Tang, Chen, Cardie, Belongie, Hariharan, and Lim]{Jia:VPT}
Menglin Jia, Luming Tang, Bor{-}Chun Chen, Claire Cardie, Serge~J. Belongie, Bharath Hariharan, and Ser{-}Nam Lim.
\newblock Visual prompt tuning.
\newblock In \emph{Proceedings of the European Conference on Computer Vision}, pages 709--727, 2022.

\bibitem[Jiang et~al.(2022)Jiang, Zhang, Huang, Ge, Ni, Lu, Zhou, Song, and Huang]{Jiang:Cross-Modal-Adapter}
Haojun Jiang, Jianke Zhang, Rui Huang, Chunjiang Ge, Zanlin Ni, Jiwen Lu, Jie Zhou, Shiji Song, and Gao Huang.
\newblock Cross-modal adapter for text-video retrieval.
\newblock \emph{arXiv preprint arXiv:2211.09623}, 2022.

\bibitem[Karthik et~al.(2022)Karthik, Mancini, and Akata]{Karthik:KG-SP}
Shyamgopal Karthik, Massimiliano Mancini, and Zeynep Akata.
\newblock {KG-SP:} knowledge guided simple primitives for open world compositional zero-shot learning.
\newblock In \emph{Proceedings of the IEEE/CVF Conference on Computer Vision and Pattern Recognition}, pages 9336--9345, 2022.

\bibitem[Khan et~al.(2023)Khan, Naeem, Gool, Pagani, Stricker, and Afzal]{Khan:CAPE}
Muhammad Gul Zain~Ali Khan, Muhammad~Ferjad Naeem, Luc~Van Gool, Alain Pagani, Didier Stricker, and Muhammad~Zeshan Afzal.
\newblock Learning attention propagation for compositional zero-shot learning.
\newblock In \emph{Proceedings of the IEEE/CVF Winter Conference on Applications of Computer Vision}, pages 3828--3837, 2023.

\bibitem[Khattak et~al.(2023)Khattak, Rasheed, Maaz, Khan, and Khan]{Khattak:MaPLe}
Muhammad~Uzair Khattak, Hanoona~Abdul Rasheed, Muhammad Maaz, Salman Khan, and Fahad~Shahbaz Khan.
\newblock {MaPLe:} multi-modal prompt learning.
\newblock In \emph{Proceedings of the IEEE/CVF Conference on Computer Vision and Pattern Recognition}, pages 19113--19122, 2023.

\bibitem[Kingma and Ba(2015)]{Kingma:Adam}
Diederik~P. Kingma and Jimmy Ba.
\newblock Adam: A method for stochastic optimization.
\newblock In \emph{Proceedings of the International Conference on Learning Representations}, 2015.

\bibitem[Lester et~al.(2021)Lester, Al{-}Rfou, and Constant]{Lester:prompt-tuning}
Brian Lester, Rami Al{-}Rfou, and Noah Constant.
\newblock The power of scale for parameter-efficient prompt tuning.
\newblock In \emph{Proceedings of the Conference on Empirical Methods in Natural Language Processing}, pages 3045--3059, 2021.

\bibitem[Li et~al.(2022)Li, Yang, Wei, Deng, and Yang]{Li:SCEN}
Xiangyu Li, Xu Yang, Kun Wei, Cheng Deng, and Muli Yang.
\newblock Siamese contrastive embedding network for compositional zero-shot learning.
\newblock In \emph{Proceedings of the IEEE/CVF Conference on Computer Vision and Pattern Recognition}, pages 9326--9335, 2022.

\bibitem[Li et~al.(2020)Li, Xu, Mao, and Lu]{Li:symmetry-and-group}
Yonglu Li, Yue Xu, Xiaohan Mao, and Cewu Lu.
\newblock Symmetry and group in attribute-object compositions.
\newblock In \emph{Proceedings of the IEEE/CVF Conference on Computer Vision and Pattern Recognition}, pages 11313--11322, 2020.

\bibitem[Liu et~al.(2021{\natexlab{a}})Liu, Ji, Fu, Du, Yang, and Tang]{Liu:P-Tuning-deep}
Xiao Liu, Kaixuan Ji, Yicheng Fu, Zhengxiao Du, Zhilin Yang, and Jie Tang.
\newblock {P-Tuning v2:} prompt tuning can be comparable to fine-tuning universally across scales and tasks.
\newblock \emph{arXiv preprint arXiv:2110.07602}, 2021{\natexlab{a}}.

\bibitem[Liu et~al.(2021{\natexlab{b}})Liu, Zheng, Du, Ding, Qian, Yang, and Tang]{Liu:prompt-tuning}
Xiao Liu, Yanan Zheng, Zhengxiao Du, Ming Ding, Yujie Qian, Zhilin Yang, and Jie Tang.
\newblock {GPT} understands, too.
\newblock \emph{arXiv preprint arXiv:2103.10385}, 2021{\natexlab{b}}.

\bibitem[Lu et~al.(2023)Lu, Liu, Guo, and Guo]{Lu:DFSP}
Xiaocheng Lu, Ziming Liu, Song Guo, and Jingcai Guo.
\newblock Decomposed soft prompt guided fusion enhancing for compositional zero-shot learning.
\newblock In \emph{Proceedings of the IEEE/CVF Conference on Computer Vision and Pattern Recognition}, pages 23560--23569, 2023.

\bibitem[Mancini et~al.(2021)Mancini, Naeem, Xian, and Akata]{Mancini:CompCos}
Massimiliano Mancini, Muhammad~Ferjad Naeem, Yongqin Xian, and Zeynep Akata.
\newblock Open world compositional zero-shot learning.
\newblock In \emph{Proceedings of the IEEE/CVF Conference on Computer Vision and Pattern Recognition}, pages 5222--5230, 2021.

\bibitem[Mancini et~al.(2022)Mancini, Naeem, Xian, and Akata]{Mancini:Co-CGE}
Massimiliano Mancini, Muhammad~Ferjad Naeem, Yongqin Xian, and Zeynep Akata.
\newblock Learning graph embeddings for open world compositional zero-shot learning.
\newblock \emph{{IEEE} Transactions on Pattern Analysis and Machine Intelligence}, 2022.

\bibitem[Misra et~al.(2017)Misra, Gupta, and Hebert]{Misra:red-wine-to-red-tomato}
Ishan Misra, Abhinav Gupta, and Martial Hebert.
\newblock From red wine to red tomato: Composition with context.
\newblock In \emph{Proceedings of the IEEE/CVF Conference on Computer Vision and Pattern Recognition}, pages 1160--1169, 2017.

\bibitem[Naeem et~al.(2021)Naeem, Xian, Tombari, and Akata]{Naeem:CGE}
Muhammad~Ferjad Naeem, Yongqin Xian, Federico Tombari, and Zeynep Akata.
\newblock Learning graph embeddings for compositional zero-shot learning.
\newblock In \emph{Proceedings of the IEEE/CVF Conference on Computer Vision and Pattern Recognition}, pages 953--962, 2021.

\bibitem[Nagarajan and Grauman(2018)]{Nagarajan:attributes-as-operators}
Tushar Nagarajan and Kristen Grauman.
\newblock Attributes as operators: Factorizing unseen attribute-object compositions.
\newblock In \emph{Proceedings of the European Conference on Computer Vision}, pages 172--190, 2018.

\bibitem[Nayak et~al.(2023)Nayak, Yu, and Bach]{Nayak:CSP}
Nihal~V. Nayak, Peilin Yu, and Stephen~H. Bach.
\newblock Learning to compose soft prompts for compositional zero-shot learning.
\newblock In \emph{Proceedings of the International Conference on Learning Representations}, 2023.

\bibitem[Paszke et~al.(2019)Paszke, Gross, Massa, Lerer, Bradbury, Chanan, Killeen, Lin, Gimelshein, Antiga, Desmaison, K{\"{o}}pf, Yang, DeVito, Raison, Tejani, Chilamkurthy, Steiner, Fang, Bai, and Chintala]{Paszke:PyTorch}
Adam Paszke, Sam Gross, Francisco Massa, Adam Lerer, James Bradbury, Gregory Chanan, Trevor Killeen, Zeming Lin, Natalia Gimelshein, Luca Antiga, Alban Desmaison, Andreas K{\"{o}}pf, Edward~Z. Yang, Zachary DeVito, Martin Raison, Alykhan Tejani, Sasank Chilamkurthy, Benoit Steiner, Lu Fang, Junjie Bai, and Soumith Chintala.
\newblock {PyTorch:} an imperative style, high-performance deep learning library.
\newblock In \emph{Proceedings of the Advances in Neural Information Processing Systems}, pages 8024--8035, 2019.

\bibitem[Pennington et~al.(2014)Pennington, Socher, and Manning]{Pennington:Glove}
Jeffrey Pennington, Richard Socher, and Christopher~D. Manning.
\newblock Glove: Global vectors for word representation.
\newblock In \emph{Proceedings of the Conference on Empirical Methods in Natural Language Processing}, pages 1532--1543, 2014.

\bibitem[Purushwalkam et~al.(2019)Purushwalkam, Nickel, Gupta, and Ranzato]{Purushwalkam:task-driven-modular-networks}
Senthil Purushwalkam, Maximilian Nickel, Abhinav Gupta, and Marc'Aurelio Ranzato.
\newblock Task-driven modular networks for zero-shot compositional learning.
\newblock In \emph{Proceedings of the IEEE/CVF International Conference on Computer Vision}, pages 3592--3601, 2019.

\bibitem[Radford et~al.(2021)Radford, Kim, Hallacy, Ramesh, Goh, Agarwal, Sastry, Askell, Mishkin, Clark, Krueger, and Sutskever]{Radford:CLIP}
Alec Radford, Jong~Wook Kim, Chris Hallacy, Aditya Ramesh, Gabriel Goh, Sandhini Agarwal, Girish Sastry, Amanda Askell, Pamela Mishkin, Jack Clark, Gretchen Krueger, and Ilya Sutskever.
\newblock Learning transferable visual models from natural language supervision.
\newblock In \emph{Proceedings of the International Conference on Machine Learning}, pages 8748--8763, 2021.

\bibitem[Sung et~al.(2021)Sung, Nair, and Raffel]{Sung:overfitting}
Yi{-}Lin Sung, Varun Nair, and Colin Raffel.
\newblock Training neural networks with fixed sparse masks.
\newblock In \emph{Proceedings of the Advances in Neural Information Processing Systems}, pages 24193--24205, 2021.

\bibitem[Vaswani et~al.(2017)Vaswani, Shazeer, Parmar, Uszkoreit, Jones, Gomez, Kaiser, and Polosukhin]{Vaswani:Transformer}
Ashish Vaswani, Noam Shazeer, Niki Parmar, Jakob Uszkoreit, Llion Jones, Aidan~N. Gomez, Lukasz Kaiser, and Illia Polosukhin.
\newblock Attention is all you need.
\newblock In \emph{Proceedings of the Advances in Neural Information Processing Systems}, pages 5998--6008, 2017.

\bibitem[Wang et~al.(2023)Wang, Liu, Jing, Chen, Liang, Wang, and Shen]{Wang:CANet}
Qingsheng Wang, Lingqiao Liu, Chenchen Jing, Hao Chen, Guoqiang Liang, Peng Wang, and Chunhua Shen.
\newblock Learning conditional attributes for compositional zero-shot learning.
\newblock In \emph{Proceedings of the IEEE/CVF Conference on Computer Vision and Pattern Recognition}, pages 11197--11206, 2023.

\bibitem[Xu et~al.(2022)Xu, Kordjamshidi, and Chai]{Xu:PromptCompVL}
Guangyue Xu, Parisa Kordjamshidi, and Joyce Chai.
\newblock Prompting large pre-trained vision-language models for compositional concept learning.
\newblock \emph{arXiv preprint arXiv:2211.05077}, 2022.

\bibitem[Yu and Grauman(2014)]{Yu:UT-Zappos}
Aron Yu and Kristen Grauman.
\newblock Fine-grained visual comparisons with local learning.
\newblock In \emph{Proceedings of the IEEE/CVF Conference on Computer Vision and Pattern Recognition}, pages 192--199, 2014.

\bibitem[Zaken et~al.(2022)Zaken, Goldberg, and Ravfogel]{Zaken:ft-bias}
Elad~Ben Zaken, Yoav Goldberg, and Shauli Ravfogel.
\newblock {BitFit:} simple parameter-efficient fine-tuning for transformer-based masked language-models.
\newblock In \emph{Proceedings of the Annual Meeting of the Association for Computational Linguistics}, pages 1--9, 2022.

\bibitem[Zang et~al.(2022)Zang, Li, Zhou, Huang, and Loy]{Zang:UPT}
Yuhang Zang, Wei Li, Kaiyang Zhou, Chen Huang, and Chen~Change Loy.
\newblock Unified vision and language prompt learning.
\newblock \emph{arXiv preprint arXiv:2210.07225}, 2022.

\bibitem[Zhang et~al.(2022)Zhang, Liang, Du, Sun, Ma, and Guo]{Tian:IVR}
Tian Zhang, Kongming Liang, Ruoyi Du, Xian Sun, Zhanyu Ma, and Jun Guo.
\newblock Learning invariant visual representations for compositional zero-shot learning.
\newblock In \emph{Proceedings of the European Conference on Computer Vision}, pages 339--355, 2022.

\bibitem[Zhang and Yang(2022)]{Zhang:multi-task-learning}
Yu Zhang and Qiang Yang.
\newblock A survey on multi-task learning.
\newblock \emph{{IEEE} Transactions on Knowledge and Data Engineering}, 34\penalty0 (12):\penalty0 5586--5609, 2022.

\bibitem[Zhou et~al.(2022)Zhou, Yang, Loy, and Liu]{Zhou:CoOp}
Kaiyang Zhou, Jingkang Yang, Chen~Change Loy, and Ziwei Liu.
\newblock Learning to prompt for vision-language models.
\newblock \emph{International Journal of Computer Vision}, pages 2337--2348, 2022.

\end{thebibliography}
}

\clearpage
\setcounter{page}{1}
\maketitlesupplementary

\appendix
\renewcommand{\thesection}{\Alph{section}}

\section{Experimental Details} \label{sec:exp_details}

In this section, we give more details about the architecture, training and evaluation for reference.

\subsection{Visual Adapter}

Here we detail how we introduce Adapter~\cite{Houlsby:Adapter} into the Transformer-based image encoder.
Specifically, we insert small learnable modules (\textit{i.e.}, adapters) after the multi-head self-attention layer and the feed-forward network inside each Transformer block.
Given an input feature $\mathbf{x} \in \mathbb{R}^{d}$, the adapter module uses a down-projection with the parameter matrix $\mathbf{W}^{down} \in \mathbb{R}^{d \times r}$ to project the feature to a lower-dimensional space specified by the bottleneck dimension $r$ ($r \ll d$), followed a nonlinear activation function $\sigma(\cdot)$, and an up-projection with the parameter matrix $\mathbf{W}^{up} \in \mathbb{R}^{r \times d}$.
Adopting a residual connection design, the overall computation of the adapter module is defined as
$$
\text{Adapter}(\mathbf{x})=\mathbf{x}+\sigma(\mathbf{x}\mathbf{W}^{down})\mathbf{W}^{up}, \nonumber
$$
where $\sigma(\cdot)$ is implemented as GELUs~\cite{Hendrycks:GELU}.
Keeping the original image encoder frozen, we only optimize the parameters of these inserted adapters during training.

\subsection{Hyperparameters}

\cref{tab:hyperparam} lists the hyperparameters that differ on each dataset and are determined with the validation performance.
For other hyperparameters, the CLIP's pre-trained word embeddings of ``a photo of'' are used to initialize all three prefixes.
For Adapter inserted into the image encoder, the bottleneck dimension $r$ is set to 64, and the dropout rate is set to 0.1.
In the cross-modal traction module, the feed-forward network first expands the dimension of the input features to $4\times$ its original value, and then shrinks it back.
The number of attention heads $h$ is 12, and the dimension of the single-head attention $d^{attn}$ is 64.
The strength parameter vector $\boldsymbol{\lambda}$ is initialized with the scalar value 0.1.
During training, we use the Adam~\cite{Kingma:Adam} optimizer and decay the learning rate of all trainable parameters by 0.5 every 5 epochs.

\begin{table}[!t]
\tablestyle{5pt}{1.0}
\setlength\tabcolsep{3pt}
\def\w{20pt} 
\scalebox{1}{
    \begin{tabular}{l|c|c|c}
    \textbf{Hyperparameter} & \textbf{MIT-States} & \textbf{UT-Zappos} & \textbf{C-GQA} \\
    \shline
    Learning rate & $10^{-4}$ & $2.5\times10^{-4}$ & $1.25\times10^{-5}$ \\
    Batch size & 64    & 64   & 64 \\
    Number of epochs & 10    & 15   & 15 \\
    Attribute dropout rate & 0.3   & 0.3   & 0 \\
    CMT dropout rate & 0.1   & 0     & 0 \\
    Weight decay & $10^{-5}$ & $10^{-5}$ & $10^{-5}$ \\
    Number of CMT layers $N$ & 3     & 2     & 2 \\
    Coefficients $\alpha^c, \alpha^s, \alpha^o$ & 1, 1, 1 & 1, 1, 1 & 1, 0.1, 0.1 \\
    \end{tabular}%
  } \vspace{-2mm}
  \caption{\textbf{Hyperparameters for different datasets.}
  }
  \label{tab:hyperparam}%
  \vspace{-1mm}
\end{table}%

\subsection{Feasibility Calibration for Open-World Setting}

Following~\cite{Mancini:Co-CGE,Nayak:CSP}, we apply the post-training feasibility calibration to filter out infeasible compositions that might be present in the open-world evaluation.
The calibration assumes that similar objects share similar states while dissimilar objects are unlikely to share states.
Therefore, given a candidate pair $c=\langle s, o \rangle$, similarities between the objects can be computed as
$$
{\rho}_{o}(s, o) = \mathop{\max}\limits_{\hat{o} \in \mathcal{O}^{se}} \frac{\phi(o) \cdot \phi(\hat{o})}{\| \phi(o) \| \| \phi(\hat{o}) \|},
$$
where 
$\mathcal{O}^{se}$ is the object set that contains those paired with the state $s$ in seen compositions.
And $\phi(\cdot)$ is an embedding function that maps the primitive to a pre-trained embedding, which is implemented with GloVe embeddings~\cite{Pennington:Glove}.
We also compute similarities between the states as ${\rho}_{s}(s, o)$ in the same way.
Next, the feasibility score for the composition $(s, o)$ can be computed by combining the two similarities with a mean pooling function $\mu$:
$$
{\rho}(s, o) = \mu({\rho}_{o}(s, o), {\rho}_{s}(s, o)).
$$

Finally, by only considering compositions above a threshold $T$, infeasible compositions can be filtered out.
And the inference of \method now becomes
$$
\hat{c} = \mathop{\arg \max}\limits_{c_{i,j} \in \mathcal{C}^{tgt}, {\rho}(s, o) > T} \left(\widetilde{p}(c_{i,j}|x)\right),
$$
where the threshold $T$ is calibrated based on the performance on the validation set.

\section{Hyperparameter Sensitivity Analysis}

In this section, we vary some key hyperparameters to examine how sensitive the proposed \method is to them.

\begin{figure}[!t]
  \centering
   \includegraphics[width=\linewidth]{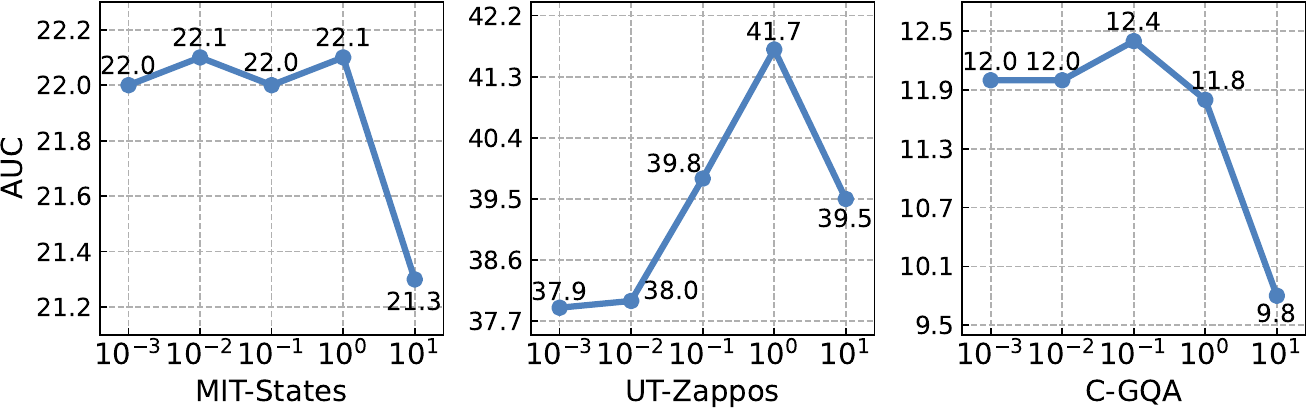}
   \vspace{-7mm}
   \caption{
   \textbf{Sensitivity analysis on loss weighting coefficients $\alpha^s$ and $\alpha^o$.}
   }
   \label{fig:prim-loss-weight}
   \vspace{-4mm}
\end{figure}

\vspace{-5mm}
\paragraph{Loss weighting coefficients $\alpha^s$ and $\alpha^o$.}
In \cref{fig:prim-loss-weight}, after fixing the weighting coefficient of the composition branch $\alpha^c$ as 1, we vary the loss coefficients on the state and object branches, \textit{i.e.}, $\alpha^s$ and $\alpha^o$.
While setting $\alpha^s$ and $\alpha^o$ as 1 can achieve the best result on MIT-States and UT-Zappos, a smaller value as 0.1 is better on C-GQA, where learning in a composable way deserves a higher priority in more complex scenarios.
Another observation is that setting $\alpha^s$ and $\alpha^o$ as 10 leads to a significant drop, 
which shows that it is detrimental to give the primitive branches a higher status than the composition branch during training.

\begin{figure}[!t]
  \centering
   \includegraphics[width=\linewidth]{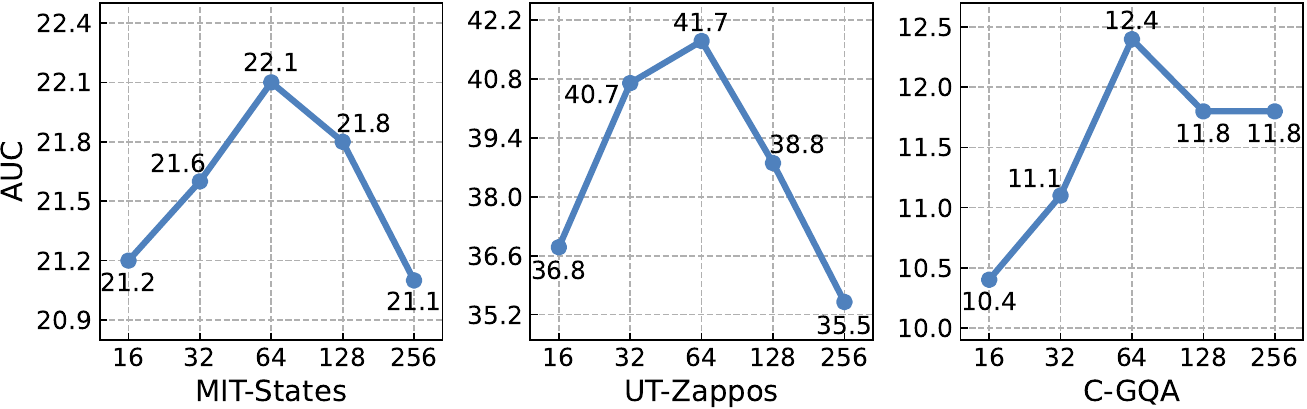}
   \vspace{-7mm}
   \caption{
   \textbf{Sensitivity analysis on Adapter bottleneck dimension $r$.}
   }
   \label{fig:adapter-dim}
   \vspace{-4mm}
\end{figure}

\vspace{-5mm}
\paragraph{Adapter bottleneck dimension $r$.}
In \cref{fig:adapter-dim}, we vary the bottleneck dimension $r$ of the introduced adapters.
For all three datasets, 64 is an optimal choice for $r$.
And a higher $r$ may cause a performance crash on smaller datasets like UT-Zappos due to over-fitting.

\begin{figure}[!t]
  \centering
   \includegraphics[width=\linewidth]{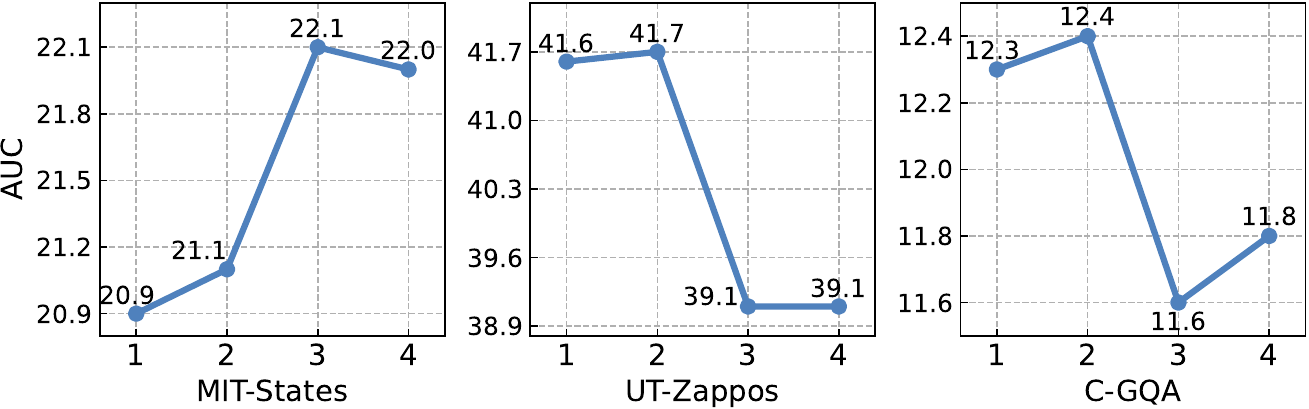}
   \vspace{-7mm}
   \caption{
   \textbf{Sensitivity analysis on the number of \CMT module layers $N$.}
   }
   \label{fig:CMT-layer}
   \vspace{-2mm}
\end{figure}

\vspace{-5mm}
\paragraph{Number of CMT module layers $N$.}
In \cref{fig:CMT-layer}, a 2-layer \CMT module is optimal for C-GQA, while setting $N$ as 3 is better for the other two datasets.
Although not affecting its leadership over baseline methods, it is observed that further deepening the module may result in a loss of performance for \method.

\begin{figure}[!t]
  \centering
   \includegraphics[width=\linewidth]{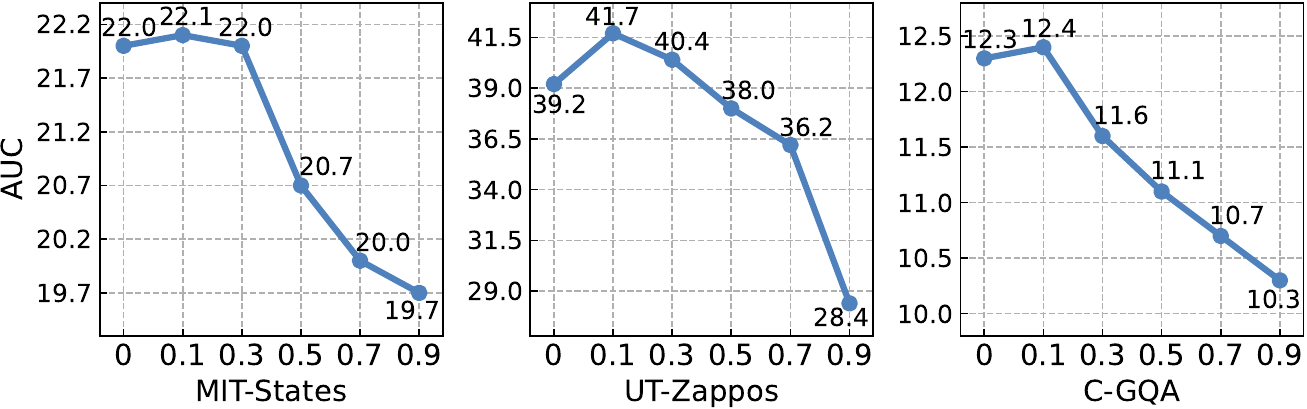}
   \vspace{-7mm}
   \caption{
   \textbf{Sensitivity analysis on initialization value of $\boldsymbol{\lambda}$.}
   }
   \label{fig:lambda-initialization}
   \vspace{-5mm}
\end{figure}

\vspace{-5mm}
\paragraph{Initialization of $\boldsymbol{\lambda}$.}
In \cref{fig:lambda-initialization}, we first vary the initialization value of the trainable parameter vector $\boldsymbol{\lambda}$, which controls the strength of the cross-modal traction.
On all three datasets, initializing $\boldsymbol{\lambda}$ with 0.1 achieves the highest AUC, and steadily increasing the initialization value leads to a continuous decline in performance.
We attribute this phenomenon to the fact that aggressive traction may destroy the cross-modal alignment already established by pre-training.
Therefore, a larger initialization value for $\boldsymbol{\lambda}$ increases the difficulty of optimization.
We also display the statistics of the trained $\boldsymbol{\lambda}$ values in \cref{tab:lambda_value}, which remain near the initial 0.1.
As a conclusion, the current initialization settings for $\boldsymbol{\lambda}$ are appropriate.

\begin{table}[t]  %
\tablestyle{5pt}{1.0}
\setlength\tabcolsep{4pt}
\def\w{20pt} 
\scalebox{1}{
    \begin{tabular}{c|c|c|c}
    \textbf{$\boldsymbol{\lambda}$ value} & \textbf{MIT-States} & \textbf{UT-Zappos} & \textbf{C-GQA} \\
    \shline
    mean  & 0.092 & 0.092 & 0.101 \\
    max   & 0.113 & 0.106 & 0.107 \\
    min   & 0.079 & 0.079 & 0.099 \\
    \end{tabular}%
    }\vspace{-3mm}
    \caption{
  \textbf{Statistics of the trained $\boldsymbol{\lambda}$.}
  }
  \label{tab:lambda_value}%
  \vspace{-2mm}
\end{table}%

\begin{table}[!t]
\tablestyle{5pt}{1.0}
\setlength\tabcolsep{3.9pt}
\def\w{20pt} 
\scalebox{1}{
    \begin{tabular}{c|cccc|cccc}
    \multirow{2}[1]{*}{\textbf{CMT in \textit{Troika}}} & \multicolumn{4}{c|}{\textbf{UT-Zappos}} & \multicolumn{4}{c}{\textbf{C-GQA}} \\
          & S     & U     & HM    & AUC   & S     & U     & HM    & AUC \\
    \shline
    \rowcolor[rgb]{ .949,  .949,  .949} $\mathbf{t} \leftarrow \mathbf{t} + \boldsymbol{\lambda} \cdot \widetilde{\mathbf{t}}$ & \textbf{66.8} & \textbf{73.8} & \textbf{54.6} & \textbf{41.7} & \textbf{41.0} & \textbf{35.7} & \textbf{29.4} & \textbf{12.4} \\
    $\mathbf{t} \leftarrow \widetilde{\mathbf{t}}$ & 64.8  & 71.9  & 48.9  & 36.1  & 40.4  & 31.6  & 28.2  & 11.1 \\
    \end{tabular}%
    } \vspace{-3mm}
    \caption{
      \textbf{Ablation on the implementation of the CMT module.}
  }
    \vspace{-2mm}
  \label{tab:resconn_ablation}%
\end{table}%

\begin{table}[!t]
\tablestyle{5pt}{1.0}
\setlength\tabcolsep{2.52pt}
\def\w{20pt} 
\scalebox{1}{
    \begin{tabular}{cc|cccc|cccc}
    \multicolumn{2}{c|}{$\boldsymbol{\lambda}$} & \multicolumn{4}{c|}{\textbf{UT-Zappos}} & \multicolumn{4}{c}{\textbf{C-GQA}} \\
    \textbf{Vectorized} & \textbf{Trainable} & S     & U     & HM    & AUC   & S     & U     & HM    & AUC \\
    \shline
    \rowcolor[rgb]{ .949,  .949,  .949} \cmark & \cmark & \textbf{66.8} & \textbf{73.8} & \textbf{54.6} & \textbf{41.7} & \textbf{41.0} & \textbf{35.7} & \textbf{29.4} & \textbf{12.4} \\
    \cmark &       & 66.2  & 73.5  & 54.2  & 41.3  & 39.8  & 33.2  & 29.1  & 11.5 \\
          & \cmark & 65.2  & 73.1  & 52.9  & 40.1  & 40.8  & 35.0  & 28.5  & 12.0 \\
    \end{tabular}%
  } \vspace{-3mm}
  \caption{\textbf{Ablation on the strength parameter $\boldsymbol{\lambda}$.} 
  }
  \label{tab:ablation-lambda}%
  \vspace{-5mm}
\end{table}%

\begin{table*}[!t]    %
\tablestyle{5pt}{1.0}
\setlength\tabcolsep{4pt}
\def\w{20pt} 
\scalebox{1}{
    \begin{tabular}{lll|cc|cc|cc}
    \multicolumn{3}{c|}{\textbf{branch}} & \multicolumn{2}{c|}{\textbf{MIT-States}} & \multicolumn{2}{c|}{\textbf{UT-Zappos}} & \multicolumn{2}{c}{\textbf{C-GQA}} \\
    \multicolumn{1}{c}{$c$} & \multicolumn{1}{c}{$s$} & \multicolumn{1}{c|}{$o$} & HM    & AUC   & HM    & AUC   & HM    & AUC \\
    \shline
    \rowcolor[rgb]{ .949,  .949,  .949} ``a photo of'' & ``a photo of'' & ``a photo of'' & \textbf{39.3} & 22.1  & 54.6  & 41.7  & 29.4  & \textbf{12.4} \\
    ``a photo of'' & ``the object is'' & ``the object is'' & 39.2  & \textbf{22.4} & \textbf{55.6} & 40.9  & \textbf{29.7} & 12.0 \\
    ``the object is'' & ``the object is'' & ``the object is'' & 38.7  & 21.7  & 52.8  & 41.0  & 28.5  & 11.5 \\
    ``a photo of'' & ``the state of the object is'' & ``the class of the object is'' & 39.1  & 22.2  & 54.5  & \textbf{42.7} & 28.7  & 11.6 \\
    ``a photo of'' & ``the state is'' & ``the class is'' & 38.8  & 21.4  & 53.2  & 38.5  & 29.6  & 12.0 \\
    \end{tabular}%
  } \vspace{-2mm}
  \caption{
  \textbf{Sensitivity analysis on initialization of prefixes.}
  }
  \label{tab:ctx-initialization}%
\end{table*}%

\vspace{-4mm}
\paragraph{Prefix initialization.}
In \cref{tab:ctx-initialization}, we report the results of trying several combinations of initialization for prompt prefixes from different branches, which confirm that \method might be marginally sensitive to the prefix initialization.
In our experiments, we have selected a simplest combination as the default initialization for convenience.

\section{Additional Ablation Study}

In this section, we add more ablation experiments to analyze the effects of each design in \method.

\begin{table*}[!t]    %
\tablestyle{5pt}{1.0}
\setlength\tabcolsep{4pt}
\def\w{20pt} 
\scalebox{1}{
    \begin{tabular}{lc|cccc|cccc|cccc}
          &       & \multicolumn{4}{c|}{\textbf{MIT-States}} & \multicolumn{4}{c|}{\textbf{UT-Zappos}} & \multicolumn{4}{c}{\textbf{C-GQA}} \\
    \textbf{Method} & \textbf{Backbone} & S     & U     & HM    & AUC   & S     & U     & HM    & AUC   & S     & U     & HM    & AUC \\
    \shline
    CLIP~\cite{Radford:CLIP}  & ViT-B/32 & 25.1  & 39.1  & 21.4  & 7.5   & 9.6   & 42.4  & 10.0  & 2.4   & 7.3   & 22.1  & 7.4   & 1.2 \\
    CSP~\cite{Nayak:CSP}   & ViT-B/32 & 36.4  & 42.5  & 28.6  & 12.4  & 57.1  & 57.3  & 39.3  & 24.2  & 30.1  & 23.4  & 19.4  & 5.7 \\
    \rowcolor[rgb]{ .949,  .949,  .949} \textbf{\method (Ours)}  & ViT-B/32 & \textbf{39.5} & \textbf{42.8} & \textbf{30.5} & \textbf{13.9} & \textbf{60.5} & \textbf{67.4} & \textbf{47.3} & \textbf{32.3} & \textbf{36.3} & \textbf{27.2} & \textbf{24.4} & \textbf{8.4} \\
    \hline
    CLIP~\cite{Radford:CLIP}  & ViT-L/14 & 30.2  & 46.0  & 26.1  & 11.0  & 15.8  & 49.1  & 15.6  & 5.0   & 7.5   & 25.0  & 8.6   & 1.4 \\
    CSP~\cite{Nayak:CSP}   & ViT-L/14 & 46.6  & 49.9  & 36.3  & 19.4  & 64.2  & 66.2  & 46.6  & 33.0  & 28.8  & 26.8  & 20.5  & 6.2 \\
    \rowcolor[rgb]{ .949,  .949,  .949} \textbf{\method (Ours)}  & ViT-L/14 & \textbf{49.0} & \textbf{53.0} & \textbf{39.3} & \textbf{22.1} & \textbf{66.8} & \textbf{73.8} & \textbf{54.6} & \textbf{41.7} & \textbf{41.0} & \textbf{35.7} & \textbf{29.4} & \textbf{12.4} \\
    \end{tabular}%
  } \vspace{-2mm}
  \caption{
  \textbf{Ablation on backbone architecture of CLIP.}
  }
  \label{tab:ablation-backbone}%
\end{table*}%

\vspace{-5mm}
\paragraph{Ablation on strength parameter $\boldsymbol{\lambda}$.}
Since an aggressive traction may destroy the cross-modal alignment already established by pre-training, the \CMT module summarizes the features with a small weight for $\widetilde{\mathbf{t}}$, avoiding its dominance.
We first prove the necessity of $\boldsymbol{\lambda}$ and the residual structure in \cref{tab:resconn_ablation}, where directly replacing $\mathbf{t}$ with $\widetilde{\mathbf{t}}$ leads to a decrease compared to the current implementation.
In \cref{tab:ablation-lambda}, we ablate $\boldsymbol{\lambda}$ with two adjustments: (1) freeze $\boldsymbol{\lambda}$ after the initialization, and (2) change the parameter vector $\boldsymbol{\lambda} \in \mathbb{R}^d$ to a trainable scalar.
We observe that each of both adjustments leads to a drop in performance, which reveals the importance of adaptively scaling the strength of the cross-modal traction performed on each dimension.

\vspace{-5mm}
\paragraph{Ablation on backbone.}
\cref{tab:ablation-backbone} compares \method with other methods when using different ViT-based CLIP backbones.
Note that only CLIP~\cite{Radford:CLIP} and CSP~\cite{Nayak:CSP} are included in the comparison, as other CLIP-based methods have not reported the results with different backbones.
We can observe that our \method consistently outperforms the compared methods, and a larger backbone leads to better performance.

\section{Additional Comparison Results}

In this section, we present a more comprehensive comparison of our \method to demonstrate its superiority.

\subsection{Efficiency Comparison with SOTA Method}

One concern may arise that, compared to the existing single-branch methods, methods following our \MP paradigm need to extract and align the multi-modal features for each branch individually, thus requiring more number of parameters and computation.
However, we point out that this can be avoided by carefully designing the efficient solutions.
For both text and vision feature extraction, the design of \method maintains the idea of parameter efficiency to minimize the number of training parameters.
Moreover, for inference, \method only needs to extract the text features for all three branches once for the whole dataset, and the decoupling of the image features occurs after the output projection, which does not require multiple forward computations by the visual encoder.
Therefore, 
when calculating the average inference time for a single sample, the increase due to the \MP paradigm is actually much lower than expected.
To demonstrate this,
we conduct an efficiency comparison with the state-of-the-art DFSP~\cite{Lu:DFSP} method on the UT-Zappos dataset, and the results are listed in \cref{tab:efficiency-comparison}.
It is observed that DFSP actually requires more trainable parameters and inference time than \method, due to its reliance on heavy cross-attention and self-attention blocks outside of the CLIP backbone for cross-modal information interaction.
As a conclusion, the proposed efficient implementations on the \MP paradigm allow our approach to outperform the state-of-the-art methods in terms of recognition accuracy without sacrificing storage and operational efficiency.

\begin{table}[th]
\tablestyle{5pt}{0.7}
\setlength\tabcolsep{3pt}
\def\w{20pt} 
\scalebox{1}{
  \begin{tabular}{l|c|c}
    \textbf{Methods} & \textbf{\#Params (M) ↓} & \textbf{Inference Time (ms) ↓} \\
    \shline \\[-1.7ex]
    \textbf{DFSP~\cite{Lu:DFSP}} & 31.81 & 18.56 \\
    \rowcolor[rgb]{ .949,  .949,  .949} \textbf{\method (Ours)} & \textbf{21.70} & \textbf{17.31} \\
  \end{tabular}
  } \vspace{-2mm}
  \caption{
  \textbf{Efficiency comparison between DFSP and \method.}
  We report the number of trainable parameters and the average inference time.
  \method is superior in terms of storage and operational efficiency.
  }
  \label{tab:efficiency-comparison}%
  \vspace{-5mm}
\end{table} 

\subsection{Comparison with Existing CZSL Methods} \label{sec:full_comparison}

\paragraph{Baselines.}
We compare \method with both CLIP-based methods~\cite{Radford:CLIP,Zhou:CoOp,Nayak:CSP,Xu:PromptCompVL,Lu:DFSP} and existing CZSL methods~\cite{Nagarajan:attributes-as-operators,Misra:red-wine-to-red-tomato,Purushwalkam:task-driven-modular-networks,Li:symmetry-and-group,Mancini:CompCos,Naeem:CGE,Mancini:Co-CGE,Li:SCEN,Anwaar:CVGAE,Karthik:KG-SP,Khan:CAPE} with a pre-trained ResNet-18~\cite{He:ResNet} backbone.
For CompCos~\cite{Mancini:CompCos} and Co-CGE~\cite{Mancini:Co-CGE}, we report the results of which version of the models according to the experimental setup, \textit{i.e.}, the closed-world version for the closed-world setting, and the open-world version for the open-world setting.

\vspace{-4mm}
\paragraph{Results.}
\cref{tab:closed-world-SOTA-full} reports the closed-world results and \cref{tab:open-world-SOTA-full} reports the open-world results.
Credit to the transferred pre-trained knowledge, we find that CLIP-based methods significantly outperform other CZSL methods in both settings.
And our proposed \method achieves the state-of-the-art performance in all cases.

\begin{table*}[!t]    %
\tablestyle{5pt}{1.0}
\setlength\tabcolsep{1pt}
\def\w{20pt} 
\scalebox{1}{
    \begin{tabular}{lcccc|cccc|cccc}
    \multirow{2}[1]{*}{\textbf{Method}} & \multicolumn{4}{c|}{\textbf{MIT-States}} & \multicolumn{4}{c|}{\textbf{UT-Zappos}} & \multicolumn{4}{c}{\textbf{C-GQA}} \\
          & S     & U     & HM     & AUC   & S     & U     & HM     & AUC   & S     & U     & HM     & AUC \\
    \shline
    AoP~\cite{Nagarajan:attributes-as-operators}   & 14.3  & 17.4  & 9.9   & 1.6   & 59.8  & 54.2  & 40.8  & 25.9  & 17.0  & 5.6   & 5.9   & 0.7 \\
    LE+~\cite{Misra:red-wine-to-red-tomato}   & 15.0  & 20.1  & 10.7  & 2.0   & 53.0  & 61.9  & 41.0  & 25.7  & 18.1  & 5.6   & 6.1   & 0.8 \\
    TMN~\cite{Purushwalkam:task-driven-modular-networks}   & 20.2  & 20.1  & 13.0  & 2.9   & 58.7  & 60.0  & 45.0  & 29.3  & 23.1  & 6.5   & 7.5   & 1.1 \\
    SymNet~\cite{Li:symmetry-and-group} & 24.2  & 25.2  & 16.1  & 3.0   & 49.8  & 57.4  & 40.4  & 23.4  & 26.8  & 10.3  & 11.0  & 2.1 \\
    CompCos~\cite{Mancini:CompCos} & 25.3  & 24.6  & 16.4  & 4.5   & 59.8  & 62.5  & 43.1  & 28.1  & 28.1  & 11.2  & 12.4  & 2.6 \\
    CGE~\cite{Naeem:CGE}   & 28.7  & 25.3  & 17.2  & 5.1   & 56.8  & 63.6  & 41.2  & 26.4  & 28.1  & 10.1  & 11.4  & 2.3 \\
    Co-CGE~\cite{Mancini:Co-CGE} & 27.8  & 25.2  & 17.5  & 5.1   & 58.2  & 63.3  & 44.1  & 29.1  & 29.3  & 11.9  & 12.7  & 2.8 \\
    SCEN~\cite{Li:SCEN}  & 29.9  & 25.2  & 18.4  & 5.3   & 63.5  & 63.1  & 47.8  & 32.0  & 28.9  & 12.1  & 12.4  & 2.9 \\
    CVGAE~\cite{Anwaar:CVGAE} & 28.5  & 25.5  & 18.2  & 5.3   & 65.0  & 62.4  & 49.8  & 34.6  & 28.2  & 11.9  & 13.9  & 2.8 \\
    CANet~\cite{Wang:CANet} & 29.0  & 26.2  & 17.9  & 5.4   & 61.0  & 66.3  & 47.3  & 33.1  & 30.0  & 13.2  & 14.5  & 3.3 \\
    CAPE~\cite{Khan:CAPE} & 30.5  & 26.2  & 19.1  & 5.8   & 60.4  & 67.4  & 45.5  & 31.3  & 32.9  & 15.6  & 16.3  & 4.2 \\
    \hline
    CLIP~\cite{Radford:CLIP}  & 30.2  & 46.0  & 26.1  & 11.0  & 15.8  & 49.1  & 15.6  & 5.0   & 7.5   & 25.0  & 8.6   & 1.4 \\
    CoOp~\cite{Zhou:CoOp}  & 34.4  & 47.6  & 29.8  & 13.5  & 52.1  & 49.3  & 34.6  & 18.8  & 20.5  & 26.8  & 17.1  & 4.4 \\
    CSP~\cite{Nayak:CSP}   & 46.6  & 49.9  & 36.3  & 19.4  & 64.2  & 66.2  & 46.6  & 33.0  & 28.8  & 26.8  & 20.5  & 6.2 \\
    PromptCompVL~\cite{Xu:PromptCompVL} & 48.5  & 47.2  & 35.3  & 18.3  & 64.4  & 64.0  & 46.1  & 32.2  & - & - & - & - \\
    DFSP(i2t)~\cite{Lu:DFSP} & 47.4  & 52.4  & 37.2  & 20.7  & 64.2  & 66.4  & 45.1  & 32.1  & 35.6  & 29.3  & 24.3  & 8.7 \\
    DFSP(BiF)~\cite{Lu:DFSP} & 47.1  & 52.8  & 37.7  & 20.8  & 63.3  & 69.2  & 47.1  & 33.5  & 36.5  & 32.0  & 26.2  & 9.9 \\
    DFSP(t2i)~\cite{Lu:DFSP} & 46.9  & 52.0  & 37.3  & 20.6  & 66.7  & 71.7  & 47.2  & 36.0  & 38.2  & 32.0  & 27.1  & 10.5 \\
    \rowcolor[rgb]{ .949,  .949,  .949} \textbf{\method (Ours)} & \textbf{49.0}\tiny{$\pm$0.4} & \textbf{53.0}\tiny{$\pm$0.2} & \textbf{39.3}\tiny{$\pm$0.2} & \textbf{22.1}\tiny{$\pm$0.1} & \textbf{66.8}\tiny{$\pm$1.1} & \textbf{73.8}\tiny{$\pm$0.6} & \textbf{54.6}\tiny{$\pm$0.5} & \textbf{41.7}\tiny{$\pm$0.7} & \textbf{41.0}\tiny{$\pm$0.2} & \textbf{35.7}\tiny{$\pm$0.3} & \textbf{29.4}\tiny{$\pm$0.2} & \textbf{12.4}\tiny{$\pm$0.1} \\
    \end{tabular}%
    } \vspace{-2mm}
  \caption{\textbf{Closed-world results.} 
  For our \method, we report the average performance on 5 random seeds with standard error.}
  \label{tab:closed-world-SOTA-full}%
\end{table*}%

\begin{table*}[!t]    %
\tablestyle{5pt}{1.0}
\setlength\tabcolsep{1pt}
\def\w{20pt} 
\scalebox{1}{
    \begin{tabular}{lcccc|cccc|cccc}
    \multirow{2}[1]{*}{\textbf{Method}} & \multicolumn{4}{c|}{\textbf{MIT-States}} & \multicolumn{4}{c|}{\textbf{UT-Zappos}} & \multicolumn{4}{c}{\textbf{C-GQA}} \\
          & S     & U     & HM     & AUC   & S     & U     & HM     & AUC   & S     & U     & HM     & AUC \\
    \shline
    AoP~\cite{Nagarajan:attributes-as-operators}   & 16.6  & 5.7   & 4.7   & 0.7   & 50.9  & 34.2  & 29.4  & 13.7  & - & - & - & - \\
    LE+~\cite{Misra:red-wine-to-red-tomato}   & 14.2  & 2.5   & 2.7   & 0.3   & 60.4  & 36.5  & 30.5  & 16.3  & 19.2  & 0.7   & 1.0   & 0.08 \\
    TMN~\cite{Purushwalkam:task-driven-modular-networks}   & 12.6  & 0.9   & 1.2   & 0.1   & 55.9  & 18.1  & 21.7  & 8.4   & - & - & - & - \\
    SymNet~\cite{Li:symmetry-and-group} & 21.4  & 7.0   & 5.8   & 0.8   & 53.3  & 44.6  & 34.5  & 18.5  & 26.7  & 2.2   & 3.3   & 0.43 \\
    CompCos~\cite{Mancini:CompCos} & 25.4  & 10.0  & 8.9   & 1.6   & 59.3  & 46.8  & 36.9  & 21.3  & 28.4  & 1.8   & 2.8   & 0.39 \\
    CGE~\cite{Naeem:CGE}   & 29.6  & 4.0   & 4.9   & 0.7   & 58.8  & 46.5  & 38.0  & 21.5  & 28.3  & 1.3   & 2.2   & 0.30 \\
    Co-CGE~\cite{Mancini:Co-CGE} & 26.4  & 10.4  & 10.1  & 2.0   & 60.1  & 44.3  & 38.1  & 21.3  & 28.7  & 1.6   & 2.6   & 0.37 \\
    KG-SP~\cite{Karthik:KG-SP} & 28.4  & 7.5   & 7.4   & 1.3   & 61.8  & 52.1  & 42.3  & 26.5  & 31.5  & 2.9   & 4.7   & 0.78 \\
    CVGAE~\cite{Anwaar:CVGAE} & 27.3  & 9.9   & 10.0  & 1.8   & 58.6  & 48.4  & 41.7  & 22.2  & 26.6  & 2.9   & 6.4   & 0.7 \\
    \hline
    CLIP~\cite{Radford:CLIP}  & 30.1  & 14.3  & 12.8  & 3.0   & 15.7  & 20.6  & 11.2  & 2.2   & 7.5   & 4.6   & 4.0   & 0.27 \\
    CoOp~\cite{Zhou:CoOp}  & 34.6  & 9.3   & 12.3  & 2.8   & 52.1  & 31.5  & 28.9  & 13.2  & 21.0  & 4.6   & 5.5   & 0.70 \\
    CSP~\cite{Nayak:CSP}   & 46.3  & 15.7  & 17.4  & 5.7   & 64.1  & 44.1  & 38.9  & 22.7  & 28.7  & 5.2   & 6.9   & 1.20 \\
    PromptCompVL~\cite{Xu:PromptCompVL} & 48.5  & 16.0  & 17.7  & 6.1   & 64.6  & 44.0  & 37.1  & 21.6  & - & - & - & - \\
    DFSP(i2t)~\cite{Lu:DFSP} & 47.2  & 18.2  & 19.1  & 6.7   & 64.3  & 53.8  & 41.2  & 26.4  & 35.6  & 6.5   & 9.0   & 1.95 \\
    DFSP(BiF)~\cite{Lu:DFSP} & 47.1  & 18.1  & 19.2  & 6.7   & 63.5  & 57.2  & 42.7  & 27.6  & 36.4  & 7.6   & 10.6  & 2.39 \\
    DFSP(t2i)~\cite{Lu:DFSP} & 47.5  & 18.5  & 19.3  & 6.8   & \textbf{66.8} & 60.0  & 44.0  & 30.3  & 38.3  & 7.2   & 10.4  & 2.40 \\
    \rowcolor[rgb]{ .949,  .949,  .949} \textbf{\method (Ours)} & \textbf{48.8}\tiny{$\pm$0.4} & \textbf{18.7}\tiny{$\pm$0.1} & \textbf{20.1}\tiny{$\pm$0.1} & \textbf{7.2}\tiny{$\pm$0.1} & 66.4\tiny{$\pm$1.0}  & \textbf{61.2}\tiny{$\pm$1.0} & \textbf{47.8}\tiny{$\pm$1.3} & \textbf{33.0}\tiny{$\pm$1.0} & \textbf{40.8}\tiny{$\pm$0.2} & \textbf{7.9}\tiny{$\pm$0.2} & \textbf{10.9}\tiny{$\pm$0.3} & \textbf{2.70}\tiny{$\pm$0.1} \\
    \end{tabular}%
  } \vspace{-2mm}
  \caption{\textbf{Open-world results.} 
  For our \method, we report the average performance on 5 random seeds with standard error.}
  \label{tab:open-world-SOTA-full}%
\end{table*}%

\end{document}